\documentclass[11pt,a4paper]{article}
\usepackage{times,latexsym}
\usepackage{url}
\usepackage[T1]{fontenc}

\usepackage[acceptedWithA]{tacl2021v1}
\usepackage{xspace,mfirstuc,tabulary}

\newif\iftaclinstructions
\taclinstructionsfalse % AUTHORS: do NOT set this to true
\iftaclinstructions
\renewcommand{\confidential}{}
\renewcommand{\anonsubtext}{(No author info supplied here, for consistency with
TACL-submission anonymization requirements)}
\newcommand{\instr}
\fi

\iftaclpubformat % this "if" is set by the choice of options

\else

\fi

\usepackage{graphicx}

\usepackage{subcaption}

\usepackage{wrapfig}

\usepackage{amsmath,amsfonts,bm}
\usepackage{amssymb,mathtools}
\usepackage{bbm}
\usepackage{nccmath}
\usepackage{multicol}
\usepackage{longtable}
\usepackage{booktabs}
\usepackage{supertabular}

\usepackage{pifont}% http://ctan.org/pkg/pifont
\newcommand{\cmark}{\text{\color{green!50!black}\ding{51}}}%
\newcommand{\xmark}{\text{\color{red!70!black}\ding{55}}}%

\newcommand{\addr}[1]{}  % Fix error with \maketitle

\usepackage{url}

\makeatletter \g@addto@macro\UrlSpecials{\do\!{\newline}}

\PassOptionsToPackage{hyphens}{url}
\PassOptionsToPackage{breaklinks}{hyperref}
\mathchardef\UrlBreakPenalty=100
\mathchardef\UrlBigBreakPenalty=100
\usepackage{breakurl}

\newcommand{\markerplus}{{\textcolor{orange!70!black}{\textbf{+}}}}
\newcommand{\markerneutral}{{\textcolor{gray!50!black}{\textbf{=}}}}
\newcommand{\markernegative}{{\textcolor{blue!70!black}{\textbf{--}}}}

\usepackage[noabbrev,capitalize,nameinlink]{cleveref}
\usepackage{fontawesome5}

\usepackage{colortbl}
\usepackage{makecell}

\usepackage{tcolorbox}
\tcbuselibrary{breakable,xparse,skins}
\definecolor{textbgcolor}{HTML}{F5F5F5}
\definecolor{promptbgcolor}{HTML}{ecf9ec}
\definecolor{workflowbgcolor}{HTML}{e6ecff}
\definecolor{slotcolor}{HTML}{cc7a00}

\newcommand{\deepseek}{\raisebox{-0.05cm}{\includegraphics[width=0.4cm]{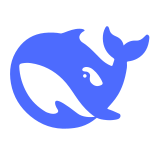}}}
\newcommand{\cohere}{\raisebox{-0.05cm}{\includegraphics[width=0.33cm]{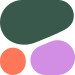}}}
\newcommand{\openai}{\raisebox{-0.05cm}{\includegraphics[width=0.35cm]{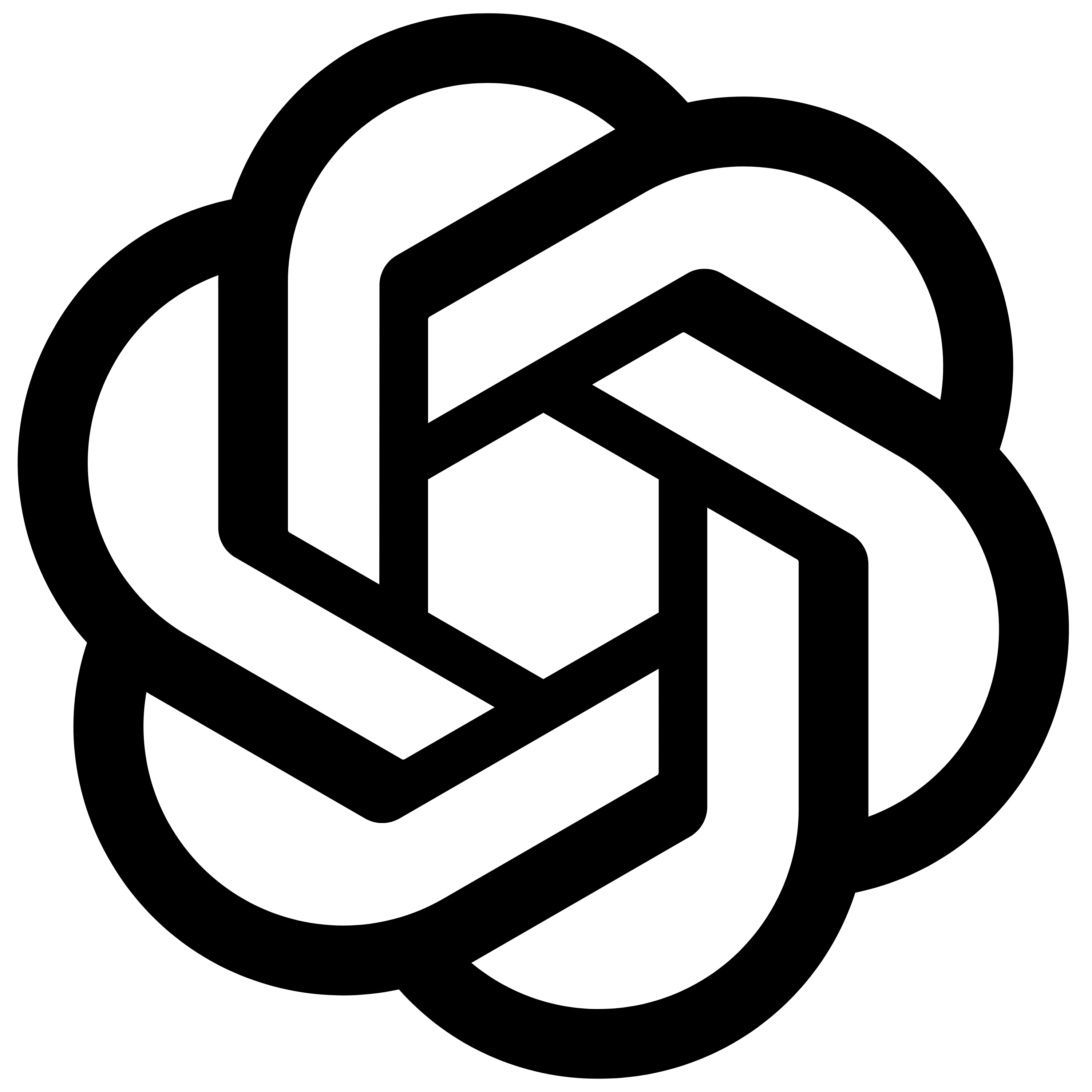}}}
\newcommand{\gemini}{\raisebox{-0.05cm}{\includegraphics[width=0.35cm]{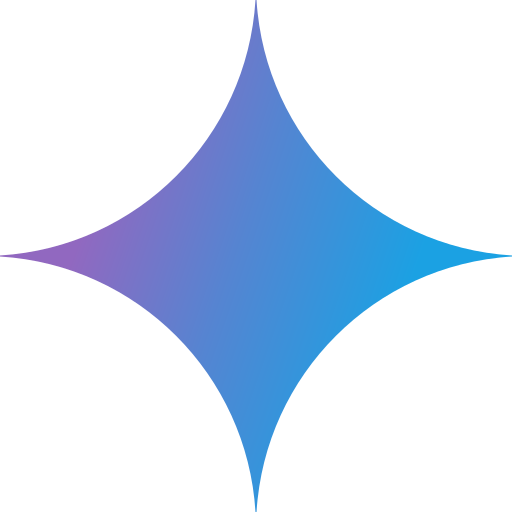}}}
\newcommand{\mistral}{\raisebox{-0.05cm}{\includegraphics[width=0.35cm]{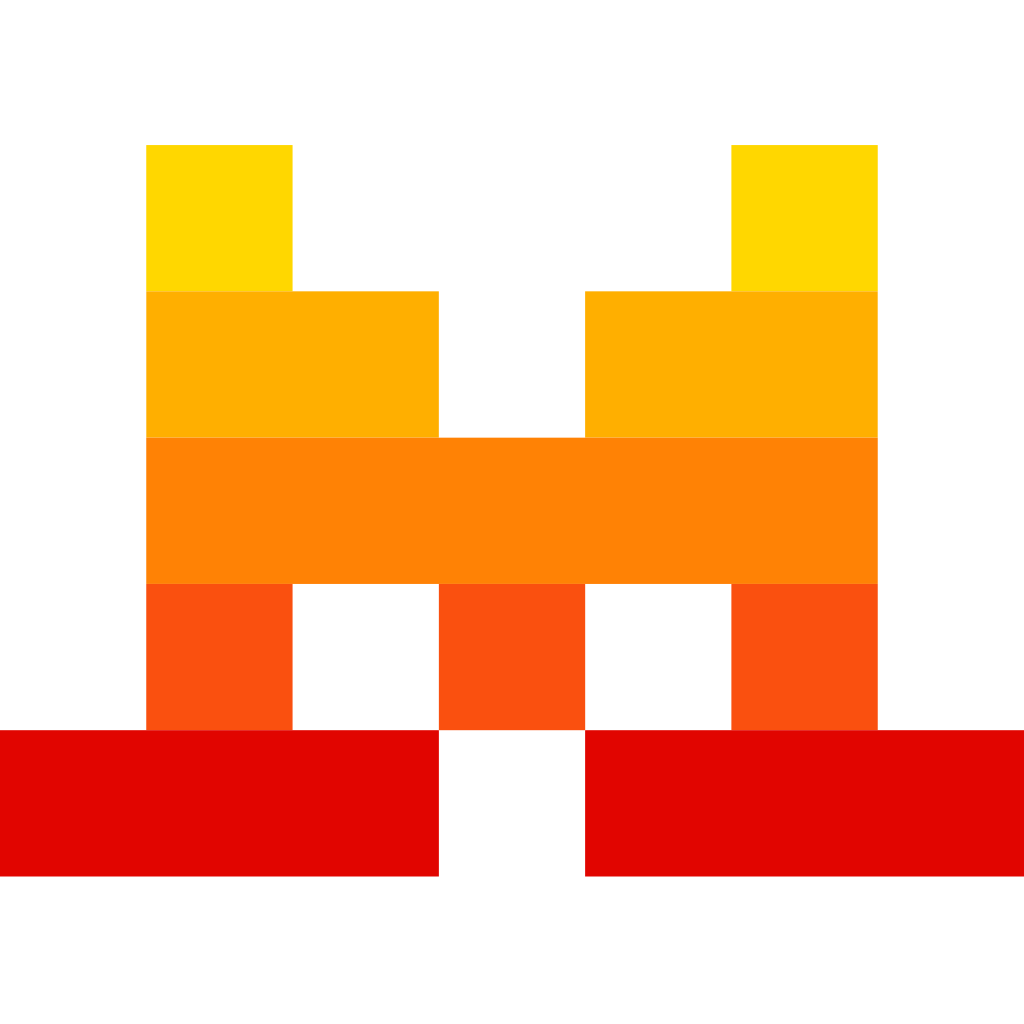}}}
\newcommand{\anthropic}{\raisebox{-0.05cm}{\includegraphics[width=0.35cm]{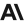}}}
\newcommand{\github}{\raisebox{-0.25cm}{\includegraphics[width=0.7cm]{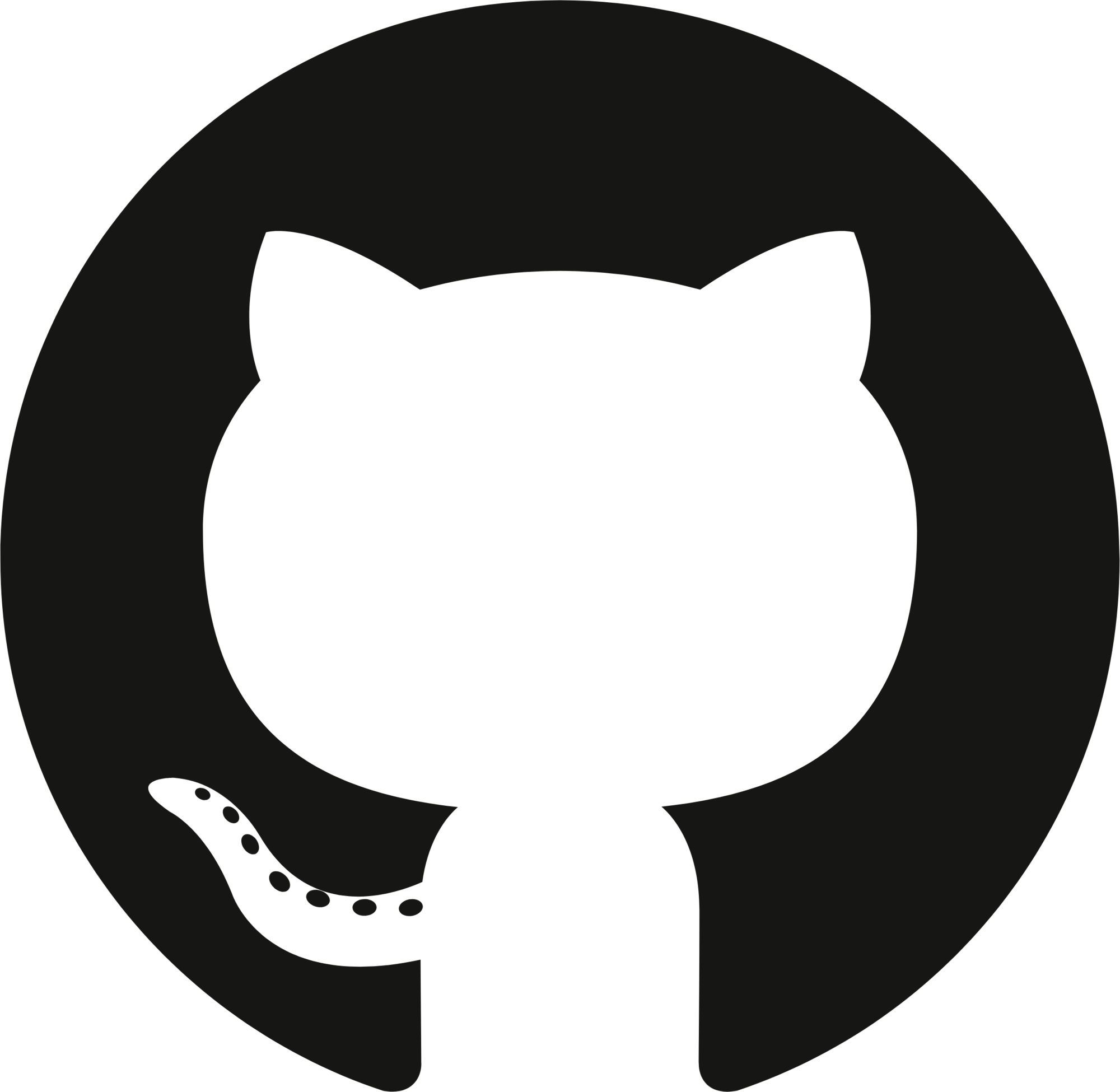}}}

\newcommand\mystrut{\rule[-1pt]{0pt}{1em}}

\newtcbox{\promptbox}[1][]{enhanced,
 box align=base,
 nobeforeafter,
 colback=orange!50!black,
 colframe=orange!50!black,
 size=small,
top=1pt,
left=1pt, bottom=1pt, right=1pt,
 boxsep=2pt,
 opacityback=0.6,enhanced jigsaw,
opacityframe=0.8,
 fontupper={\color{white}\tiny\sffamily\mystrut},
 #1}
 \newtcbox{\sftbox}[1][]{enhanced,
 box align=base,
 nobeforeafter,
 colback=green!50!black,
 colframe=green!50!black,
 size=small,
top=1pt,
left=1pt, bottom=1pt, right=1pt,
 boxsep=2pt,
 opacityback=0.6,enhanced jigsaw,
opacityframe=0.8,
  fontupper={\color{white}\tiny\sffamily\mystrut},
 #1}

 \newtcbox{\rlbox}[1][]{enhanced,
 box align=base,
 nobeforeafter,
 colback=blue!80!gray,
 colframe=blue!80!gray,
 size=small,
top=1pt,
left=1pt, bottom=1pt, right=1pt,
opacityback=0.6,enhanced jigsaw,
opacityframe=0.8,
 boxsep=2pt,
  fontupper={\color{white}\tiny\sffamily\mystrut},
 #1}

 \newtcbox{\verbalbox}[1][]{enhanced,
 box align=base,
 nobeforeafter,
 colback=red!70!black,
 colframe=red!70!black,
 size=small,
top=1pt,
left=1pt, bottom=1pt, right=1pt,
opacityback=0.6,enhanced jigsaw,
opacityframe=0.8,
 boxsep=2pt,
  fontupper={\color{white}\tiny\sffamily\mystrut},
 #1}

 \newtcbox{\numbox}[1][]{enhanced,
 box align=base,
 nobeforeafter,
 colback=cyan!80!black,
 colframe=cyan!80!black,
 size=small,
top=1pt,
left=1pt, bottom=1pt, right=1pt,
opacityback=0.6,enhanced jigsaw,
opacityframe=0.8,
 boxsep=2pt,
  fontupper={\color{white}\tiny\sffamily\mystrut},
 #1}

\newcommand{\promptvanilla}{\promptbox{Prompting: Vanilla}}
\newcommand{\promptcot}{\promptbox{Prompting: Chain-of-thought}}
\newcommand{\promptself}{\promptbox{Prompting: Self-assessment}}
\newcommand{\promptreasoning}{\promptbox{Prompting: Reasoning}}
\newcommand{\sft}{\sftbox{Supervised Finetuning}}
\newcommand{\sftdistil}{\sftbox{Supervised Finetuning: Knowledge Distillation}}
\newcommand{\sftother}{\sftbox{Supervised Finetuning: Other}}
\newcommand{\rldpo}{\rlbox{Reinforcement Learning: DPO}}
\newcommand{\rlppo}{\rlbox{Reinforcement Learning: PPO}}

\newcommand{\verballikert}{\verbalbox{Verbal: Likert Scale}}
\newcommand{\verbalyesno}{\verbalbox{Verbal: Yes / No}}
\newcommand{\verbalmarkers}{\verbalbox{Verbal: Epistemic Marker}}
\newcommand{\verbalfluent}{\verbalbox{Verbal: Fluent}}
\newcommand{\verbalnumhedges}{\verbalbox{Verbal: Number of hedges}}
\newcommand{\numzerohundred}{\numbox{Numerical: $0-100$ / $0-1$}}

\newcommand{\numzeroten}{\numbox{Numerical: $0-10$}}
\newcommand{\numzeronine}{\numbox{Numerical: $0-9$}}
\newcommand{\numci}{\numbox{Numerical: Confidence Interval}}
\newcommand{\numoneten}{\numbox{Numerical: $1-10$}}
\newcommand{\numonehundred}{\numbox{Numerical: $1-100$}}

\definecolor{findingscolor}{HTML}{8DD3A4} 

\title{Anthropomimetic Uncertainty:\\ What Verbalized Uncertainty in Language Models is Missing}
\author{Dennis Ulmer\textsuperscript{\faBicycle}, \hspace{.5em} Alexandra Lorson\textsuperscript{\faComments},\hspace{.5em} Ivan Titov\textsuperscript{\faBicycle, \faChessRook },\hspace{.5em} Christian Hardmeier\textsuperscript{\faCompass, \faFlag}\\
      \addr \textsuperscript{\faBicycle } ILLC, University of Amsterdam \textsuperscript{\faComments} CLCG, University of Groningen\\
       \textsuperscript{\faChessRook} ILCC, University of Edinburgh \textsuperscript{\faCompass} IT University of Copenhagen\\\ \textsuperscript{\faFlag} Pioneer Centre for Artificial Intelligence\\
       \small{
        \textbf{Correspondence:} \href{mailto:dennis.ulmer@mailbox.org}{dennis.ulmer@mailbox.org}
        }}

\begin{document}
\maketitle

\begin{abstract}
    Human users increasingly communicate with large language models (LLMs), but LLMs suffer from frequent overconfidence in their output, even when its accuracy is questionable, which undermines their trustworthiness and perceived legitimacy.
    Therefore, there is a need for language models to signal their confidence in order to reap the benefits of human-machine collaboration and mitigate potential harms.
    \emph{Verbalized uncertainty} is the expression of confidence with linguistic means, an approach that integrates perfectly into language-based interfaces.
    Most recent research in natural language processing (NLP) overlooks the nuances surrounding human uncertainty communication and the biases that influence the communication of and with machines.
    We argue for \emph{anthropomimetic uncertainty}, the principle that intuitive and trustworthy uncertainty communication requires a degree of imitation of human linguistic behaviors.
    %linguistic authenticity and personalization to the user.
    We present a thorough overview of the research in human uncertainty communication, survey ongoing research in NLP, and perform additional analyses to demonstrate so-far underexplored biases in verbalized uncertainty.
    We conclude by pointing out unique factors in human-machine uncertainty and outlining future research directions towards implementing anthropomimetic uncertainty.\\

    \vspace{-0.2cm}
    \begin{tabular}{rl}
        {\github} & \href{https://github.com/Kaleidophon/anthropomimetic uncertainty}{\makecell[cl]{\footnotesize \texttt{kaleidophon/}\\\footnotesize  \texttt{anthropomimetic-uncertainty}}} \\
    \end{tabular}
    %\github\hspace{0.15cm}\href{https://github.com/Kaleidophon/anthropomimetic uncertainty}{\footnotesize \texttt{kaleidophon/\\\hspace{0.4cm}anthropomimetic-uncertainty}}
\end{abstract}

\begin{figure}[htb]
    \centering
    \includegraphics[width=0.85\columnwidth]{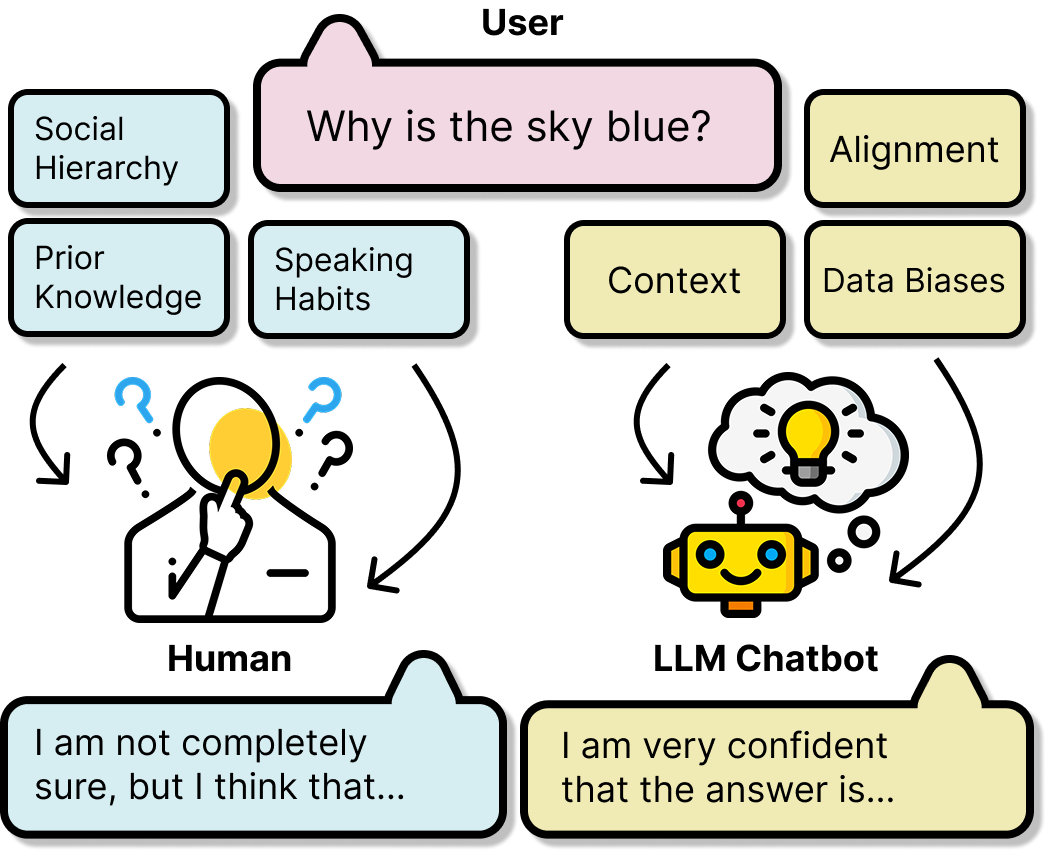}
    \caption{Humans consider various factors when expressing uncertainty, including prior knowledge or the relationship to the speaker. LLMs are biased by their training data, conversational context, and alignment for helpfulness, leading to overconfident and untrustworthy responses.}\label{fig:intro-figure}
\end{figure}

\section{Introduction}
% https://www.cjr.org/tow_center/we-compared-eight-ai-search-engines-theyre-all-bad-at-citing-news.php
In March 2025, the Columbia Journalism Review tested whether different artificial intelligence (AI) chatbots could accurately identify and cite news articles \citep{jazwinska2025ai}.
Given a snippet, the chatbots were tasked to identify information such as article date, headline and publisher.
Not only did many chatbots give incorrect but plausible-looking answers, but they also stated them with complete confidence: 
For instance, ChatGPT answered incorrectly 67\% of times, hedged its answers using phrases such as ``it appears'' or ``it's possible'' in only 7.5\% of responses, and never declined to answer any question.
%Thus, while chatbots remain a powerful and personalized tool, their replies often remain misleading or wrong.
This investigation gives a glimpse into the risks of LLMs assistants; while they remain a powerful tool, their replies can often be misleading or wrong. %which is reshaping societies right now. 
Potential exposure is high---for instance, \citet{eloundou2024gpts} estimate that LLMs are relevant to $14 \%$ of tasks across 1016 analyzed jobs.
Simultaneously, adoption is increasing \citep{humlum2024adoption} with LLMs being used for everyday tasks \citep{wang2024understanding}, and freelance work being  substituted \citep{teutloff2025winners}.
%considered more and more for education \citep{kasneci2023chatgpt}, and used 
Humans have a tendency to perceive LLMs as human \citep{miller2019explanation, devrio2025taxonomy}.
This is problematic since LLMs, unlike humans, tend to hallucinate like in the example above, i.e.\@\ produce plausible-looking but wrong information  \citep{guerreiro2023hallucination, ye2024cognitive} with unwarranted confidence \cite{krause2023confidently, yona2024can, xiong2024can}.
This behavior resembles strategies used by humans for deception, and reveals a deeper challenge: 
the usual cognitive mechanisms against being accidentally or intentionally misinformed are less effective in human–machine interactions. 
Users can only assess the response, but not the reasoning process or communicative intent. 
Assessing content is difficult as people increasingly treat LLMs as sources of information \cite{Gude2025}, and thus, lack the knowledge to evaluate its plausibility.
This risks a potentially irreversible loss of trust  \citep{dhuliawala2023diachronic, zhou024relying}, as well as negative outcomes.
%Having LLMs express their confidence or uncertainty using words---a line of research referred to as \emph{verbalized uncertainty} \citep{lin2022teaching, mielke2022reducing}---can help to potentially mitigate negative side-effects from wrong predictions.
%Unfortunately, language models usually express themselves confidently \citep{zhou024relying}, and their own confidence assessments tend to be inaccurate \citep{krause2023confidently, yona2024can, xiong2024can}.
Yet, consistent uncertainty communication can help build extrinsic trust \cite{jacovi2021formalizing}, 
but despite growing attention in NLP as \emph{verbalized uncertainty} \citep{lin2022teaching, mielke2022reducing}, its basis in human communication and the implications for AI remain underexplored:
As depicted in \cref{fig:intro-figure}, our work discusses how human uncertainty is influenced by factors such as prior knowledge, speaking habits and social hierarchy, while LLM uncertainty is shaped by conversational context, data biases, and its alignment procedure.\\

%\extodo{Integrate into previous paragraph}Problems arise when users treat human–machine interaction as analogous to human–human communication. For instance, LLMs frequently produce false information \cite[e.g.,][]{jazwinska2025ai}, more than what is expected of a presumably cooperative speaker. Compounding this issue, LLMs tend to present false information with unwarranted confidence \cite{krause2023confidently, yona2024can, xiong2024can}, mimicking a strategy used by uncooperative human speakers, but without any underlying intent to deceive. This reveals a deeper challenge: the standard mechanisms of epistemic vigilance are less effective in human–machine interactions. Users can only assess the model's output, with no access to a speaker's reasoning processes or communicative intent. Relying on the assessment of content is difficult as users increasingly treat LLMs as sources of information \cite{Gude2025}, and thus, lack the necessary background knowledge to evaluate its plausibility.

%is a line of research referred to as \emph{verbalized uncertainty} \citep{lin2022teaching, mielke2022reducing}.

%Having LLMs express their confidence or uncertainty using words is a line of research referred to as \emph{verbalized uncertainty} \citep{lin2022teaching, mielke2022reducing}.

\paragraph{Contributions.}
In this work we introduce \emph{anthropomimetic uncertainty}: 
Grounding verbalized uncertainty in norms of human uncertainty communication in order to improve reliability, trustworthiness and personalization.
To this end, \cref{sec:linguistic-perspectives} first gives a detailed account of the nuances in human uncertainty communication by drawing from the linguistics literature.
After a short introduction into uncertainty quantification in NLP in \cref{sec:background}, \cref{sec:research-nlp} contrasts the linguistic account with the current state of verbalized uncertainty research in NLP:
By summarizing and annotating the existing literature, we demonstrate that most works still rely on insufficient prompting methods and solely use numerical expressions.
This is not enough: We show through frequency analysis in different corpora that the occurrences and linguistic markers is highly dependent on the domain of text. 
We analyze the uncertainty communication in the generations of five different commercial LLMs, empirically showing that their use of uncertainty expressions is heavily influenced by conversational context and domain. 
We demonstrate that they exhibit extreme variability and inconsistent calibration, with calibration errors of up to $44 \%$. 
When using strengtheners and weakeners, they also show the whole range from perfect usage to anti-correlation when compared to the actual answer correctness.
To build a path forward, we discuss unique challenges arising in human-machine pragmatics that are linked to uncertainty communication, such as anthropomorphization and sycophancy.
%Lastly, we propose a list of actionable research directions towards anthropomimetic uncertainty, like choosing the right way to communicate uncertainty, a call to rethink calibration, personalization of uncertainty, and its role with other factors in building user trust. 
Lastly, we propose a list of actionable research directions towards anthropomimetic uncertainty, such as: Rethinking calibration, personalization of uncertainty, expansion to more languages, and considering other factors alongside uncertainty that build user trust.

\section{Human Communication of Uncertainty}\label{sec:linguistic-perspectives}

Although predictive uncertainty can be represented in many ways \citep{hoaraumeasures}, humans are intuitively able to express and interpret uncertainty verbally.
This section first provides a better understanding of the norms and mechanisms in human uncertainty communication.
We distinguish two \emph{registers} of verbal uncertainty, namely numerical and verbal expressions, where words such as \emph{certain}, \emph{might} or \emph{maybe} are called epistemic markers.\\

Human communication is understood as the intentional exchange of relevant propositions \cite{Grice1975, Sperber1995}---a proposition being the core meaning or claim of a sentence, being evaluated as true or false. 
Both speaker and addressee invest communicative effort, guided by cooperative principles to be informative, truthful, relevant, and clear \citep{Grice1975}. 
As a result, the addressee gains true and relevant information, while the speaker achieves their goal of eliciting an effect or response in the addressee \cite{Sperber2010}. 
A key aspect of cooperative behavior is acknowledging uncertainty when the speaker is unsure about a proposition. 
While a speaker expresses some level of confidence in  ``\textit{Mo will pass the exam}'', they can express uncertainty via verbal (e.g., \textit{It is possible that Mo will pass the exam}) and numerical expressions (e.g., \textit{There is a 40\% chance Mo will pass the exam}).
Thus, comprehension involves a tentative stance of trust \cite{Holton1994}, such that the addressee can revise their beliefs based on the speaker's contribution, which is advantageous given that communication is so ubiquitous \cite{Sperber2001, Bergstrom2006}. 
However, the interests of speaker and addressee rarely align perfectly: 
while the addressee expects true and relevant information, the speaker may instead choose the message that is most likely to produce the desired effect, regardless of its truth---for instance by under- or overstating one's confidence. 
This divergence makes the addressee exercise caution in order to avoid being misled, engaging in \emph{epistemic vigilance}, a cognitive capacity that helps them guard against both accidental and intentional misinformation \citep{Sperber2010}. 
According to \citeauthor{Sperber2010}, the credibility of information depends primarily on two factors: the perceived reliability of the source and the plausibility of the content. 
Thus, comprehension should neither be equated with belief nor adopting a skeptical stance.
In the following section, we examine how uncertainty is expressed, how it can lead to miscommunication due to imprecision, why speakers may strategically adjust their confidence, and what this means for human-machine interaction.

\subsection{Verbal uncertainty expressions}\label{sec:verbal}

One way of communicating uncertainty is through verbal expressions involving epistemic markers that range from factive verbs (\textit{notice}, \textit{think}, \textit{believe}), modal verbs (\textit{must}, \textit{might}), to adverbs (\textit{probably}, \textit{certainly}), which can form increasingly complex expressions (\textit{a slight yet plausible chance}). 
Their meaning is often captured by mapping them onto probabilities, intervals, or thresholds \citep{Lichtenstein1967, BeythMarom1982, wesson2009verbal, Lassiter2010, vogel2022interpretation}.\footnote{E.g.\@\ quantitative semantic approaches pose that any verbal expression is true given a threshold in $[0,1)$ if the probability of a described event exceeds it \cite{Lassiter2010, Lassiter2017, Moss2015, Swanson2006, Yalcin2007, Yalcin2010}.} 
However, their meaning is inherently imprecise and context-dependent, lacking consistent interpretation across speakers and situations.
Empirical studies suggest that interlocutors interpret \citep{Mosteller1990} and produce \citep{Tiel2022} the epistemic marker \textit{possible} to refer to probabilities between 17\%--89\%, with an average of 42\%. 
There are various factors shown to influence the production and interpretation of verbal uncertainty expressions, including event severity and base rates: 
participants interpreted verbal expressions as indicating higher likelihoods for more frequent events compared to rarer ones \cite{Wallsten1986}. 
When base rates were controlled, these expressions also signaled greater likelihood for events with more severe consequences relative to less severe ones \cite{Weber1990, Bonnefon2006, Harris2011}. 
%\textit{possible} was found to be interpreted as indicating higher numerical probabilities when used to describe a more severe condition (e.g., deafness) compared to a less severe but equally prevalent condition (e.g., insomnia) \citep{Bonnefon2006}
Despite their imprecision, verbal uncertainty expressions help speakers approximate their confidence, as they can rarely compute precise probabilities. 

%This same imprecision can benefit human–machine communication by allowing models to express uncertainty without suggesting a precise probability -- something that could falsely imply the model has accurate insight into the likelihood that its output is true.

\subsection{(Inter-)personal Motivations}\label{sec:interper-ego-mot}

Epistemic markers add complexity to uncertainty communication, as they can be used to either strengthen or hedge utterances \citep[e.g.,][]{Fraser1975, Holmes1982}. 
Strengthening \textit{You misunderstood this paragraph} involves high-certainty markers (in bold), e.g., \textit{I \textbf{am sure} that you misunderstood this paragraph} \cite{Hyland1991, Hyland2005}; and hedging low-certainty markers, e.g., \textit{You \textbf{might} have misunderstood this paragraph} \citep{Holtgraves2016}. 
A speaker may strengthen a statement for persuasion, or hedge for politeness \citep{Brown1978}.
Whether speakers engage in polite behavior depends on social hierarchies or relative power dynamics: 
speakers are more likely to use hedging with someone of higher social status, such as a manager, teacher, or parent  \citep{Brown1978, Holtgraves2010, Holtgraves2014, Holtgraves2016, Juanchich2015}. 
Speakers are more likely to hedge when discussing serious consequences such as poor financial decisions \citep{Juanchich2015}. 
This is taken into account by addressees: \textit{possible} was found to be interpreted as more likely when used for a more severe condition (e.g., deafness) than a less severe but equally prevalent condition (e.g., insomnia) \citep{Bonnefon2006}.
%Strengthening or hedging utterances relates to the speaker's level of commitment and the degree of accountability they assume. When making an assertion---e.g., \textit{The deadline is tomorrow}---the speaker commits to the truth of that proposition \citep{Brandom1983}. The stronger the expression of commitment, such as \textit{I am sure that the deadline is tomorrow}, the more the speaker endorses the claim. This increased commitment also raises their accountability, making them more vulnerable to scrutiny if the information proves incorrect. Work by \citet{Ochs1976} suggests that speakers hedge their statements to avoid accountability and unpleasant social consequences that may result from their communicative contribution. This strategy can help preserve credibility, as overstating confidence may backfire: studies have shown that while confident speakers initially appear more persuasive, once their message proves inaccurate, addressees adjust their trust and rely more on speakers who initially expressed uncertainty \cite{Tenney2007, Tenney2008, Vullioud2017}. Thus, hedged statements allow speakers to signal uncertainty without appearing uninformed, fostering more resilient trust when errors occur. 
%Strengthening or hedging utterances 
Whether speakers strengthen or hedge their statements is also shaped by how much commitment and accountability they are willing to assume, balancing persuasive intent with the potential social consequences of being wrong. 
The stronger the expression of commitment (e.g., \textit{I am sure that the deadline is tomorrow}), the more the speaker endorses the claim. 
This in turn increases scrutiny if wrong, while hedging can help speakers avoid negative social consequences and preserve credibility \citep{Ochs1976}. 
Research shows that although confident speakers seem persuasive at first, their credibility declines when proven inaccurate, whereas hedged statements foster more resilient trust \citep{Tenney2007, Tenney2008, Vullioud2017}.
%As a note, facework and managing one's accountability can equally well explain hedging: high-power speakers may hedge to avoid being inappropriately face-threatening or because they require higher confidence to make strong claims under conditions of increased accountability.
Independent of motivation, miscommunication may occur if the addressee fails to take into account the speaker's perspective in choosing a verbal uncertainty expression or by ignoring the speaker-addressee dynamics. 
Importantly, LLMs might be intuitively perceived to follow similar motivations, but we argue later in \cref{sec:data-biases,sec:context} how this can lead to uncertainty miscommunication through biases in their training.
%While we know that models frequently present false information confidently, in contrast to human-human communication, the ever increasing usage of LLMs suggests that this loss of trust may be short-lived or outweighed by their perceived utility. However, some research suggests that users do loose their trust when models repeatedly produce false information [REF]\todo.
%Recent research highlights that natural language uncertainty can mitigate overreliance on LLMs: \citet{kim2024m} found that first-person uncertainty expressions (e.g., I'm not sure) reduce user confidence and agreement while improving accuracy, likely by curbing overreliance on incorrect outputs. 

\subsection{Numerical Expressions}\label{sec:num-expr}

Given the complexity of verbal expressions, numerical formats may appear more precise and less error-prone. 
These include percentages, fractions, decimals, and ratios, which may convey relatively precise information (e.g., \textit{a 65\% probability}) or relatively imprecise estimates in the form of ranges (e.g., \textit{60\% to 70\%}). 
Previous studies found a tendency that speakers prefer verbal and addressees numerical expressions \citep{Wallsten1993, Erev1990}. 
However, preference for either has been shown to vary depending on the domain and context. 
For example, in medicine \cite{Juanchich2020b} and intelligence analysis \cite{Barnes2016}, contexts requiring a high precision and where miscommunication can come with dire consequences, speakers tend to prefer numerical expressions,\footnote{
    It should be noted however that standardized usages of verbal expressions exist in these fields, see e.g.\@\  \citet{europeancommission2009, ministryofdefence2023}.
} 
 with more serious consequences prompting a stronger preference \cite{Juanchich2020b}.\\
Although numerical expressions of uncertainty may seem precise, numbers like 65 in \textit{there is a 65\% chance of rain} often represent approximations, as multiples of 5 are treated as round numbers \cite{Ruud2014, Cummins2015, Jansen2001, Krifka2007}. 
Speakers frequently round for simplicity (e.g., saying 50 instead of 47), and such numbers are typically understood as approximate. 
In contrast, non-round numbers like 52 imply precision; referring to 52 as 50 is usually acceptable, but the reverse can seem misleading \cite{Lasersohn1999}. 
%People use round numbers approximately for several reasons. When precision is unnecessary for the communicative goal, round numbers allow speakers to remain informative without being overly exact, aligning with Grice's maxim to be only as precise as is necessary \cite{Grice1975, Gibbs2008, VanHenst2004}. In such cases, unnecessary precision may even be perceived as pedantic or socially inappropriate \cite{Beltrama2023, Cotterill2007, Lin2013, McCarthy2006}. Additionally, speakers often use round numbers when the exact value is unknown, such as estimating a crowd size, to signal reduced epistemic commitment \cite{Ruud2014}, as round numbers are generally interpreted approximately unless explicitly marked (e.g., with \textit{exactly}) \cite{Krifka2007, Lasersohn1999}. 
Speakers often use round numbers when precision is unnecessary, aligning with Grice's maxim to be only as precise as needed: round numbers allow speakers to remain informative without being overly exact \cite{Grice1975, Gibbs2008, VanHenst2004}. Unnecessary precision, in contrast, may be perceived as pedantic or socially inappropriate \cite{Beltrama2023, Cotterill2007, Lin2013, McCarthy2006}. 
Speakers also use round numbers when the exact value is unknown to signal reduced commitment \cite{Krifka2007, Lasersohn1999}.
%as they are generally interpreted as approximate unless explicitly marked (e.g., with \textit{exactly}) \cite{Krifka2007, Lasersohn1999}.
%\footnote{Jansen and Pollmann (2001) define roundness in terms of four key properties -- 10-ness, 2-ness, 2.5-ness, and 5-ness -- which describe numbers that yield small integers when divided by a power-of-10 multiple of 1, 2, 2.5, or 5; the more of these properties a number has, the rounder and more frequent it tends to be.} 
%Round numbers are cognitively salient (Van der Herbst \& Sperber, 2004) and frequently used approximately, especially at higher magnitudes where numerical cognition becomes less precise (Cheyette \& Piantadosi, 2019; McCrink \& Wynn, 2007; Xu \& Spelke, 2000). 
%Thus, round number may resemble more directly imprecise ways of communicating about quantities, including vague quantifiers such as \textit{many} and noun expressions such as \textit{masses of} (Cummins, 2015; Moxey, 2023; Channell, 1994). 
This suggests that even when uncertainty is expressed numerically, the communicative use of round numbers introduces a layer of imprecision that somewhat parallels verbal uncertainty expressions.
Numerical expressions are therefore not the silver bullet for uncertainty communication: 
Especially in machines they can imply precision when it is not warranted, and might be misinterpreted by listeners with low numeracy skills \citep{zikmund2007validation, galesic2010statistical}.

\begin{tcolorbox}[width=\columnwidth,enhanced, colback=white, coltitle=white, top=6pt, colframe=findingscolor, left=-2pt, title=\centering{\textcolor{black}{Section Findings}}, ] 
\small
\begin{itemize}
    \setlength\itemsep{0.05em}
    \item Uncertainty communication is not just influenced by the likelihood of an event, but also social hierarchies, communicative preferences, and other speaker-listener dynamics.
    \item Numerical expressions can imply precision but are often rounded by speakers; verbal expressions are imprecise, which can be an advantage for topics for which precise probabilities are not available.
\end{itemize}
\end{tcolorbox}

\section{From Human to Human-Machine}\label{sec:human-to-machine}

%Related to human-to-machine interaction, verbal uncertainty expressions align with what \citet{devrio2025taxonomy} described as expressions of intelligence, self-assessment, perspective, (dis)agreeableness, and/or embodiment in their taxonomy of linguistic features that contribute to the anthropomorphism of language technology. For example, \textit{I think that X} expresses the involvement of a thinking and reflecting intelligence with a perspective, and \textit{I see that X} suggests bodily features and functions. %Thus, while including verbal uncertainty expressions in large language models could help to highlight the models uncertainty in propositions, their inclusion could increase anthropomorphism, as was found by XIJ. This may have good (reF) and bad consequences (ref)

\cref{sec:linguistic-perspectives} displayed an extensive picture of the nuances of human uncertainty communication.
%, where speakers don't just  communicate their lack of knowledge, but also consider the social hierarchy between people, the type of person they address, and express politeness or commitment.
Contemporary language technology also mimics a conversational setting, making many of these considerations become blurry.
%Can I trust the responses by an LLM, especially when they include expressions of uncertainty?
%Does a chat assistant follow human communication behaviors or does it deviate in unexpected ways? 
In this section, we explore the current state of research and highlight some caveats to the uncertainty communication of LLMs, before also exploring some unique notions.
%To answer these questions, this section is structured as follows:
%After briefly giving some background information in \cref{sec:background}, \cref{sec:research-nlp} gives an overview over existing research in verbalized uncertainty in NLP.
%Afterwards, we demonstrate that uncertainty expressions by LLMs are not grounded in self-awareness or agency, but rather influenced by their training procedure (\cref{sec:data-biases}) and the conversational context \cref{sec:context}).
%Lastly, we outline other novel influences on the pragmatics of human-machine uncertainty communication in \cref{sec:pragmatic-considerations}.

\subsection{Background}\label{sec:background}

%To understand the place of verbalized uncertainty research in NLP, we first introduce some core concepts. 
Uncertainty is a multi-faceted concept with many different definitions and interpretations based on the context.
We start from the definition of uncertainty by \citet{baan2023uncertainty}, who define the concept as the lack or limit of knowledge of an agent interacting with the (or a) world. 
This definition lends itself well to the case of language models interacting with human users (or other language models), even if these interactions occur purely in the form of generating language.
%\citet{baan2023uncertainty} define uncertainty as the lack or limit of knowledge of an agent interacting with the  world.
%This definition lends itself well to the case of language models interacting with human users (or other language models), even if these interactions happen purely in the form of generating language.
In machine learning, the most prominent paradigms to \emph{quantify} this uncertainty draw from frequentist and Bayesian statistics.
%Quantifying predictive uncertainty is usually performed through the lens of statistical modeling, namely \emph{frequentist} and \emph{Bayesian} statistics.
In the frequentist view, probabilities correspond to the relative frequency of an event, under repetitions of an experiment \citep{willink2012disentangling}.
Uncertainty can then be quantified through  \emph{confidence},\footnote{Another way is the set / interval size in conformal prediction \citep{angelopoulos2023conformal, campos2024conformal}.} where some score $\hat{p} \in [0, 1]$ is obtained (corresponding to the model's probability of the most likely outcome).
The quality of an uncertainty estimate is assessed through \emph{calibration} \citep{guo2017calibration}.
Given a prediction $\hat{y}$, a model is calibrated if $p(y = \hat{y} \mid \hat{p}) = \hat{p}$, i.e.\@\ if we observe $\hat{p} = 0.45$, then $\hat{y}$ should be correct 45 out of 100 times on average under repetitions of the experiment.
In contrast, the Bayesian perspective views probabilities as degrees of belief in the likelihood of an event  \citep{good197146656}, and uncertainty is expressed in posterior distributions over predicted quantities. 
Bayesian methods also decompose uncertainty into \emph{data} (aleatoric) and epistemic (model) uncertainty \citep{shafer1976mathematical, shafer1978non, der2009aleatory, hullermeier2021aleatoric}, where data uncertainty refers to unresolvable ambiguity, randomness or noise.
In contrast, model uncertainty concerns the lack of knowledge about the best model, which can usually be corrected by increasing the amount of training data.\footnote{
    However, practical decompositions are often unreliable \citep{mucsanyi2024benchmarking}, do not account for all sources of uncertainty \citep{gruber2023sources}, and both uncertainties can appear in (ir-)reducible forms \citep{baan2023uncertainty, ulmer2024uncertainty}.
    %---data uncertainty might be reducible when for instance improving ambiguities in annotation guidelines, and model uncertainty might be irreducible when due to a misspecified hypothesis class.
}

\paragraph{Uncertainty in Modeling Language.}
NLP models are also faced with uncertainty contained in human language.
Language is often fundamentally underspecified \citep{pustejovsky2017semantics}, context-dependent, imprecise and ambiguous \citep{kennedy2011ambiguity}, creating uncertainty in the input data and modeling: 
For instance, saying \emph{she didn’t make it} does not specify whether she didn't manage to complete something (like a race) or didn't create something (like a homemade vase), \emph{I saw the person with the telescope} is ambiguous about which person possesses the instrument, and \emph{glass} can refer to container or content depending on the context (\emph{I drank the glass} vs.\@\ \emph{I filled up the glass}).
%Nevertheless, as we as humans only possess incomplete knowledge about the world around us, language becomes a way to communicate our perceived state of reality.
% In this work we therefore aim to draw parallels between how the uncertainty in data and models can be expressed verbally by NLP models, similar to the uncertainty communication of humans.
Compared to pure quantification of uncertainty, verbalized uncertainty presents a completely new paradigm: 
It can serve two different roles, namely the communication of uncertainty alone (when it was estimated through a separate method), or the joint quantification \& communication (when the model is prompted for its uncertainty estimate in the form of a natural language response).
In this process, two different notions arise: 
Uncertainty about the validity of a response, as well as about its verbalization. 
While \citet{baan2023uncertainty} discuss uncertainty in the output generation process, this distinction has, to the best of our knowledge, not been discussed for verbalized uncertainty.
The methods discussed here focus on the former sense.
%namely separate communication or joint quantification and communication of uncertainty.
%In the former case, we directly ask the model to verbalize its uncertainty, while in the latter case we obtain uncertainty estimates from another source, and then train the model to verbalize them.

\subsection{Verbalized Uncertainty in NLP}\label{sec:research-nlp}

\begin{figure*}[htb!]
    %[htb!]
    %\begin{subfigure}[b]{0.328\textwidth}
    %    \centering 
    %    \includegraphics[width=0.985\linewidth]{img/%publications.png}
    %    \vspace{1cm}
    %    \caption{
    %    Publications per year and category.
    %    }\label{fig:publications}
    %\end{subfigure}%
    %\hfill
     %\begin{minipage}{0.35\linewidth}
     \begin{subfigure}[b]{0.485\textwidth}
        \centering
        %\vspace{-1cm}
        \includegraphics[width=0.96\linewidth]{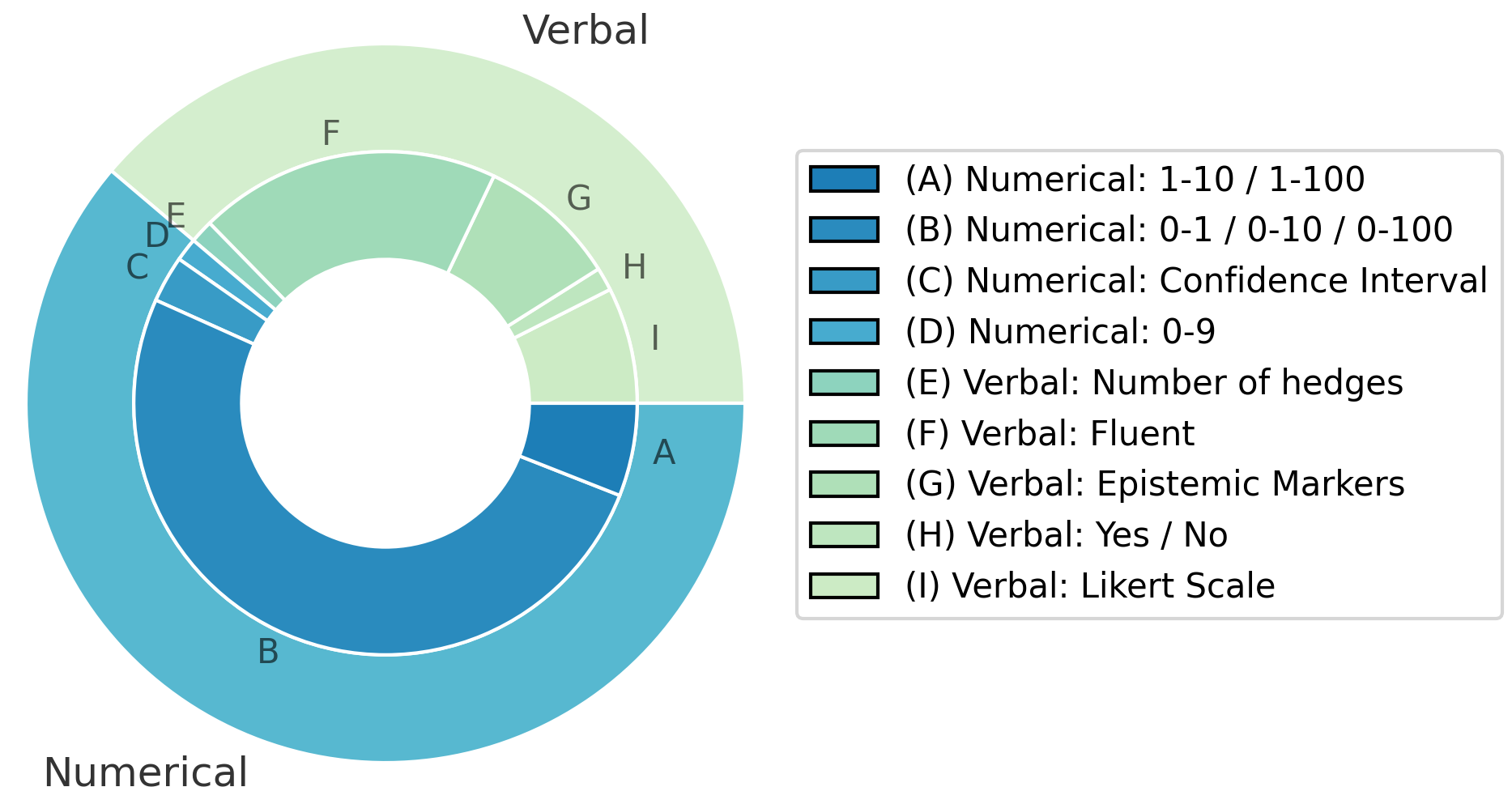}
        %\vspace{-0.42cm}
        \caption{Registers of verbalization.}\label{subfig:registers}
     \end{subfigure}%
     %\end{minipage}
     \hfill
    % \begin{minipage}{0.45\linewidth}
     \begin{subfigure}[b]{0.505\textwidth}
        \centering
        %\vspace{-1.4cm}
        \includegraphics[width=0.96\linewidth]{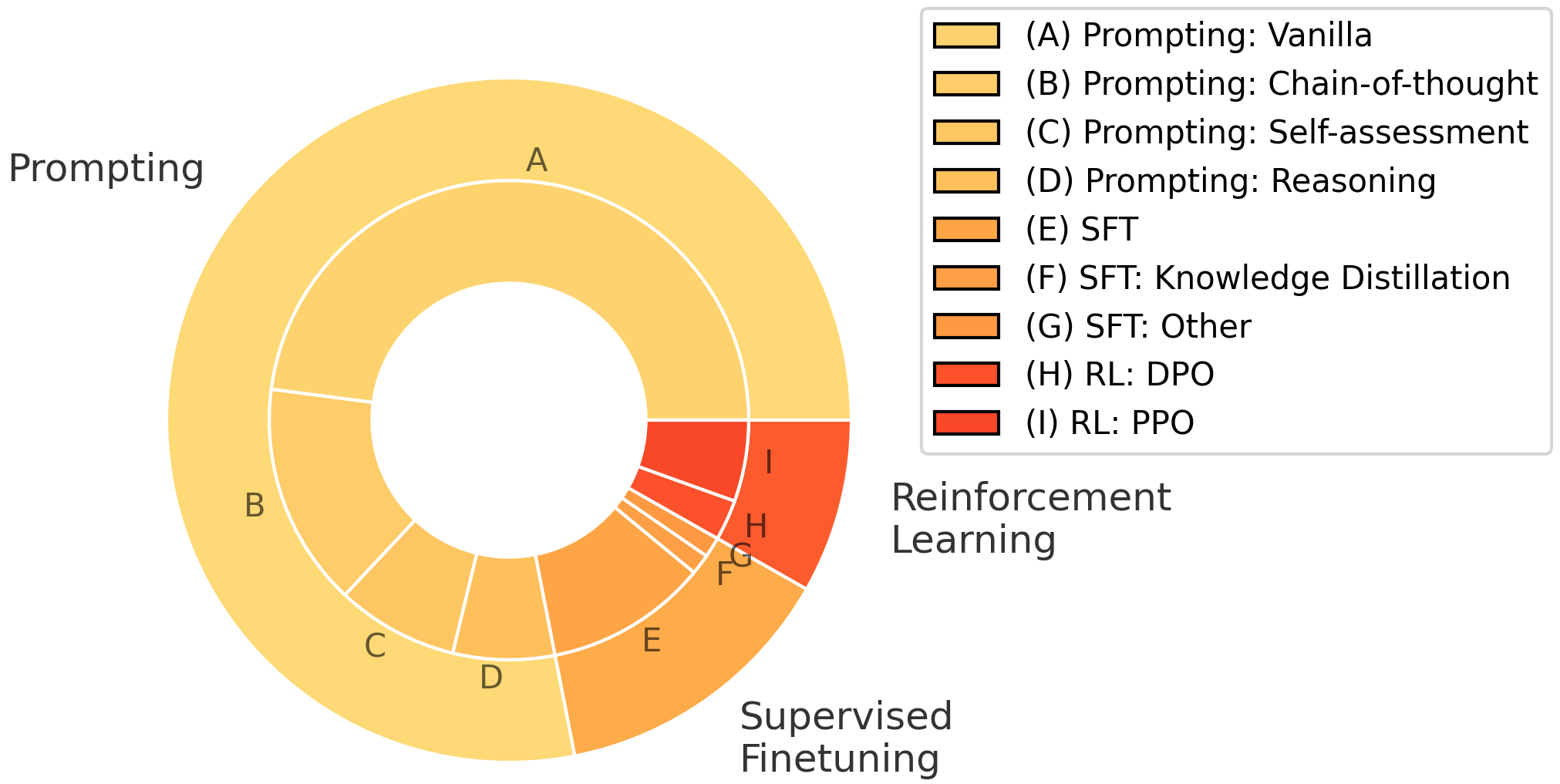}
        \caption{Method of elicitation.}\label{subfig:eliciations}
     \end{subfigure}
     %\end{minipage}
        \caption{Analysis of surveyed papers. 
        %(a) Overview over available publications about verbalized uncertainty (June 2025). 
        (a) Registers of verbalized uncertainty found in papers. (b) Ways in which uncertainty is elicited, i.e.\@\ either through prompting or by finetuning the model.}
        \label{fig:publications-analysis}
\end{figure*}

We collected publications to analyze the state of verbalized uncertainty research,\footnote{We considered *ACL venues as well as TMLR, ICML, ICLR, NeurIPS, AISTATS, and ArXiv in our review. All relevant papers can be found with annotations in \cref{app:surveyed-papers}.} totaling $53$ relevant publications since 2022, which marks, to the best of our knowledge, the earliest papers on the topic.
Publications were selected by searching for keywords such as \emph{verbal(ized) / linguistic uncertainty}, \emph{uncertainty communication} etc. (see \cref{tab:literature-search-keywords} in the appendix for details) for language models and then manually filtering by relevance, excluding survey papers.
Notably, the topic has enjoyed increasing interest, spurred by the works of \citet{lin2022teaching, mielke2022reducing}.
\citeauthor{lin2022teaching} ask for percentage scores and then finetune the model's calibration.
\citeauthor{mielke2022reducing} use an auxiliary predictor that predicts the confidence of a target model, which is mapped to a set of control tokens for different levels of uncertainty to guide generation.
%(corresponding to certainty, uncertainty and refusal), that are used on a specially finetuned target model to generate produce linguistic expressions of uncertainty in its answer.
This lays the groundwork for the main ideas in the literature.
% Approaches to verbalized uncertainty in NLP can roughly be categorized across the following dimensions:
The first prompts models to express uncertainty directly, while the second tries to generate confidence targets that are used for finetuning, such that a model automatically verbalizes its uncertainty. 
%The following paragraphs roughly follow the categories from \cref{fig:publications}.
While many of the analyzed works involve applications, we focus here on methodological works.

\paragraph{Prompting Techniques.} 
%We start with the first here.
When prompting, the goal is to elicit verbalization in the form of either verbal or numerical expressions, including either expressions that are usually ordered on a scale such as \emph{almost no chance}, \emph{probably} and \emph{almost certain}, which are mapped back to values in $[0, 1]$ to evaluate calibration,\footnote{This mapping is defined either heuristically or using the human ratings obtained from studies such as \citet{Lichtenstein1967, BeythMarom1982, wesson2009verbal, fagen2019perception, vogel2022interpretation}.} or mapping to percentages or scales (e.g.\@\ $1$ to $10$).
In this vein, works have explored the robustness of verbalized uncertainty to prompt wording \citep{yang2024verbalized} and language \citep{krause2023confidently}, different prompting techniques to improve calibration \citep{tian2023just, xiong2024can, liu2025metafaith}, verbalizing the model's answer distribution  \citep{yona2024can, wang2024calibrating}, and connections to other confidences measures \citep{ni2024large, kumar2024confidence} and applications in reasoning \citep{yoon2025reasoning, podolak2025read,  zeng2025thinking} and long-form generations \citep{zhang2024atomic}.

\paragraph{Finetuning Techniques.}
%Finetuning techniques include both supervised finetuning and 
For \emph{supervised finetuning (SFT)}, the pipeline first consists of using a secondary technique to create confidence targets, for instance token probabilities \citep{duan2023shifting}, the probability of the true / false token after asking for the truthfulness of the answer \citep{kadavath2022language}, or self-consistency \citep{wang2022self}. 
By mapping them to epistemic markers (or leaving them as numerical expressions) which are added to the answer, the model can be finetuned directly. 
Next to \citet{lin2022teaching}, this approach is also followed by \citet{liu2024can, chaudhry2024finetuning, hager2025uncertainty, band2024linguistic, xu2024sayself, eikema2025teaching, tao2025can}, only \citet{li2025conftuner} finetune token probabilities directly.
% where confidence targets are obtained from reweighted token probabilities (SAR; \citealp{duan2023shifting}), the probability of the true / false token after asking the model for the truthfulness of its answer ($p(\text{True})$; \citealp{kadavath2022language}), or the self-consistency of multiple sampled responses for the same answer \citep{wang2022self}, respectively, with the last approach also being utilized by \citet{band2024linguistic, xu2024sayself}.
The other approach relies on \emph{reinforcement learning}, with some techniques based on reinforcement learning from human feedback (RLHF; \citealp{christiano2017deep, ziegler2019fine, ouyang2022training}).
Whereas RLHF creates preference pairs from human feedback, here these are usually bootstrapped from existing data. 
%Where usually pairs would be constructed by asking human annotators to prefer responses that reflect certain values such as helpfulness and lack of toxicity, here preference would be constructed by preferring calibrated verbalizations of uncertainty.
LACIE \citep{stengel2024lacie} for instance uses a speaker and a listener model, where preferences pairs are constructed based on speaker uncertainty, listener acceptance and correctness, before finetuning the speaker model with direct preference optimization (DPO; \citealp{rafailov2023direct}). 
%The speaker model expresses uncertainty about its answer, while the lisener decides whether to accept the confident or uncertain answer. 
%Information about acceptance and correctness is then used to construct preference pairs, which are used to finetune the speaker model with direct preference optimization (DPO; \citealp{rafailov2023direct}). 
A related listener model approach is used by \citet{band2024linguistic}, albeit employing explicit rewards and optimization via proximal policy optimization (PPO; \citealp{schulman2017proximal}).
PPO is also applied by \citet{tao2024trust} to reward answer quality and correspondence between confidence and response quality.
\citet{xu2024sayself} combine SFT with PPO to optimize a reward for alignment between confidence and correctness.
\citet{leng2024taming} modify DPO and PPO by adapting the training objective for numerical uncertainty expressions and finetuning the reward model.
%With the recent rise of reasoning models (e.g.\@ \citealp{jaech2024openai, guo2025deepseek, yang2025qwen3}), 
\citet{yang2024can} leverage reasoning chains and verbal confidences from a teacher model to finetune a student model. %complementing \citeauthor{yoon2025reasoning}'s \citeyear{yoon2025reasoning} aforementioned work that relies solely on prompting.

\paragraph{Analysis of Current Research.} 
%We further analyze the current research landscape in \cref{fig:publications-analysis}:
%In \cref{subfig:registers}, we visualize the registers of verbalization, and find that the majority of methods ($61 \%$) still rely on numerical means, i.e.\@ some range of $0-1$ or $1-100$. 
We analyze the implementation of verbalized uncertainty next: 
\Cref{subfig:registers} shows that $61\%$ of methods use numerical registers, typically ranging from $0$–$1$ or $1$–$100$.
These expressions are easy to estimate with external methods, simple to extract from generations and straightforward to evaluate. 
Using verbal markers however requires decisions about their selection and their mapping to probabilities, and only a minority ($19 \%$) considers fluent verbalization, where epistemic markers are integrated into the answer in a natural  way.
Similarly, the majority of the works ($78 \%$) shown in \cref{subfig:eliciations} relies on prompting to elicit verbalized uncertainty.
Another large portion ($14 \%$) relies on supervised finetuning, with only a few recent works \citep{stengel2024lacie, band2024linguistic, xu2024sayself, leng2024taming} using reinforcement learning. 
%This speaks to the unrealized potential of this line of research: 
%Fluent, finetuning-based approaches might be more technically challenging, but is likely to reap the benefits from emulating human uncertainty communication.
%On the flip side, simple prompting approaches suffer from several biases, as we demonstrate next.
This highlights the contrast between the richness and nuance of human uncertainty expressions and current verbalized uncertainty in NLP: 
while finetuning-based approaches for fluency are more technically challenging, they can better emulate the complexity  of human uncertainty communication than adding a numerical expression or verbal marker to a response.
In addition, we show next how prompting methods, especially for numerical uncertainty, are prone to different biases.
%where training targets are obtained through self-consistency \citep{wang2022self}, SAR \citep{duan2023shifting} or other means. 

\subsection{Data Biases}\label{sec:data-biases}

\begin{figure*}[htb]
    \centering
    \begin{subfigure}[t]{0.58\textwidth}
        \centering
        \includegraphics[width=0.96\linewidth]{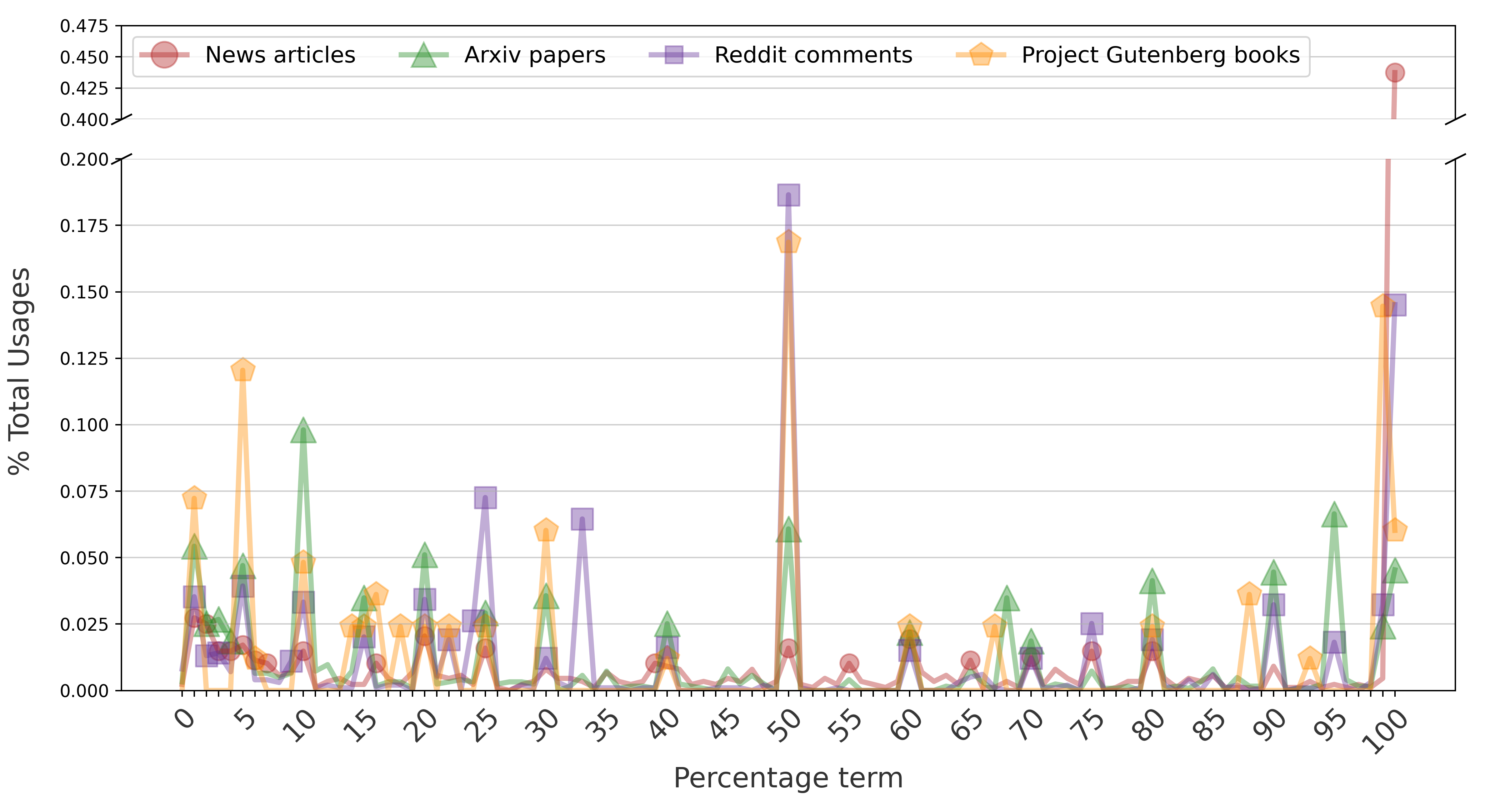}
        \caption{Distribution of number terms.}
        \label{subfig:confidence-distribution}
    \end{subfigure}
    \hfill
    \begin{subfigure}[t]{0.38\textwidth}
        \centering
        \includegraphics[width=0.96\linewidth]{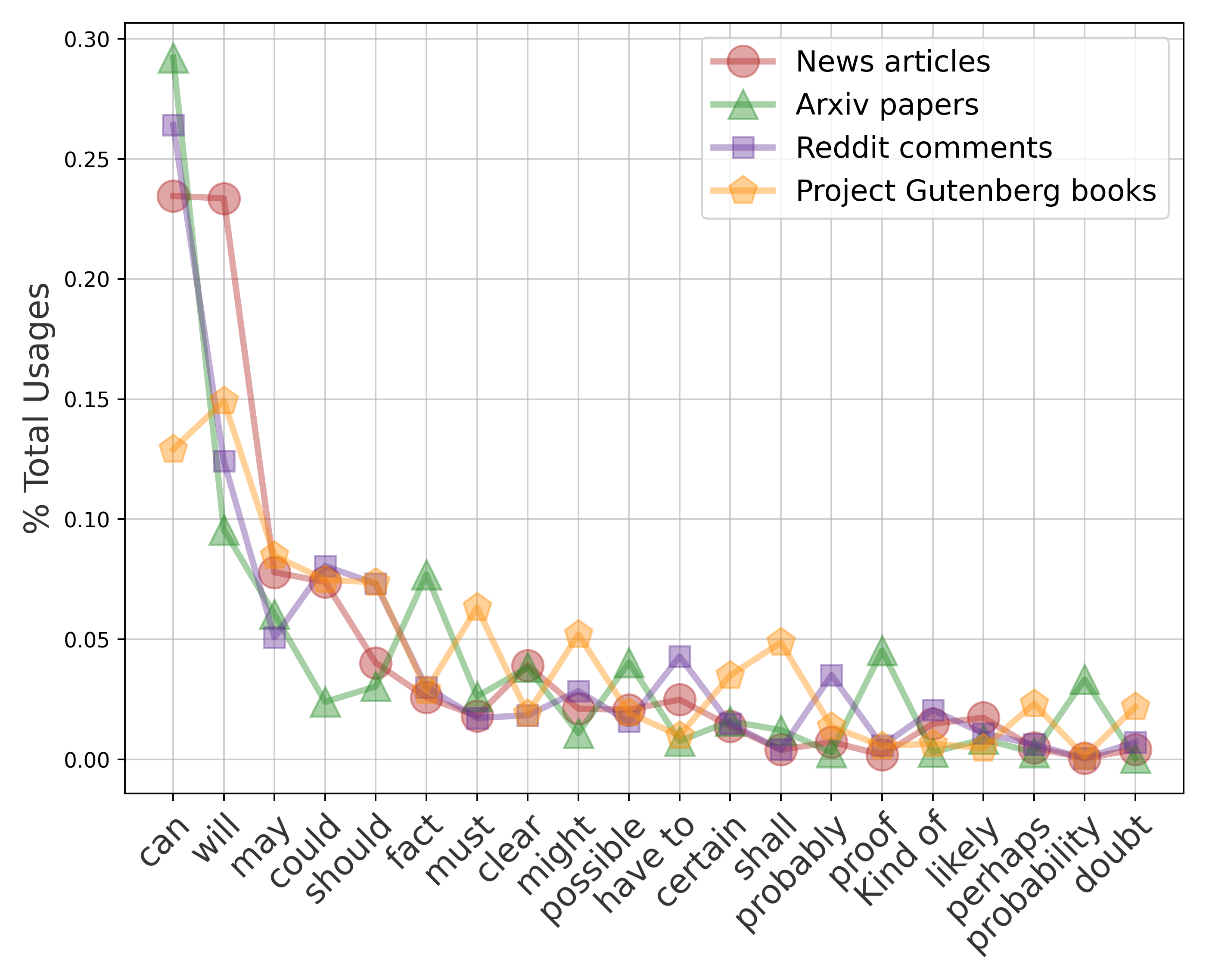}
        %\vspace{0.05cm}
        \caption{Distribution of epistemic markers.}
        \label{subfig:marker-distribution}
    \end{subfigure}
    \caption{Visualization of data-inherent biases. We plot the distribution of percentage terms in different domains (left) and the distributions of the most frequent epistemic markers (right).} %Depending on the domain, certain percentage terms and markers are mentioned much more often, potentially influencing the probability with which an LLM might generate them in a similar context.}
    \label{fig:data-biases}
\end{figure*}

%A larger number of works have shown that prompting for  verbalized uncertainty usually produces uncalibrated confidence estimates \citep{krause2023confidently, ulmer2024calibrating, xiong2024can, wang2024calibrating, groot2024overconfidence, tanneru2024quantifying, kumar2024confidence, zhang2024calibrating, sun2025large}, which we now show to be likely rooted in training data biases.
%As the generation of LLMs are largely shaped by the data they are trained on, it is intuitive to investigate the data for possible biases.

Many works have shown that prompting produces uncalibrated verbalized confidences \citep{ulmer2024calibrating, xiong2024can, wang2024calibrating, groot2024overconfidence, tanneru2024quantifying, kumar2024confidence, zhang2024calibrating, sun2025large, tao2025revisiting, bakman2025reconsidering}, which we argue through frequency analysis to be caused by training data biases. 

\paragraph{Bias in Pretraining Data.} 
For percentage-valued confidence scores, \citet{zhou2023navigating} investigate the distribution of numbers in The Pile \cite{gao2020pile}, a large pre-training corpus, and find a very skewed distribution of number terms. 
Outside of an NLP context, \citet{woodin2024large} see similar patterns for human number usage in the (much smaller) British national corpus \citep{bnc2007british}, finding that smaller and rounded numbers (to the next 5 or 10) appear more often.
%, and that the diversity of used numbers is higher in informal contexts.
%While we do not have access to the pretraining data of many commercial LLMs and open models like OLMo 2 \citep{olmo20242} publish their data without domain information, 
Since we don't have access to either pretraining data of LLMs or corresponding domain information,
we instead source datasets through Huggingface's \texttt{datasets} platform \citep{lhoest2021datasets}, where we compare the distribution of number terms and epistemic markers.\footnote{
An overview over datasets is given in \cref{table:pretraining-datasets}, and markers of interest are listed in \cref{table:markers} in the appendix.}\footnote{
The term ``domain'' is loosely defined in NLP and roughly refers to the genre or category of text. We refer to \citet{barrett2024can} for an exploration of this question.
}
\cref{subfig:confidence-distribution} shows an unequal distribution of number terms per domain: frequently appearing in quotes of people expressing absolute certainty, the number 100 appears exceedingly often in newspaper articles; the number 50 is much more frequently observed in books, and Reddit comments than for instance in ArXiv papers.\footnote{
    We also observed numerical terms above $100$ (e.g.\@\ \emph{I am 1000 \% certain}), but omitted them due to their relative rarity.
}
We conduct a similar analysis for markers by collecting them from prior work \citep{zhou2023navigating, zhou2024rel}, expanding the list through LLM-based chat assistants, verifying it through a linguistics expert, and running regex searches to count markers.\footnote{We lowercase markers and target strings and subtract cases in which another marker is a substring of the current markers (e.g.\@ ``im\emph{possible}''), as well as negated forms from a manually curated list available in our open-source repository.}
The same applies to epistemic markers in \cref{subfig:marker-distribution}, which vary across domains.
This suggests that LLMs will likely be biased in their verbalization not (only) by their uncertainty, but by their training data and context as well.

%where ``shall'' dominates due to its high frequency in the legal domain, but is almost absent from Reddit comments.%\footnote{We note that in the legal context, ``shall'' is usually not used to express uncertainty, but obligation \citep{garner2001dictionary}.}

\begin{figure}[htb]
    \centering
    \includegraphics[width=0.99\linewidth]{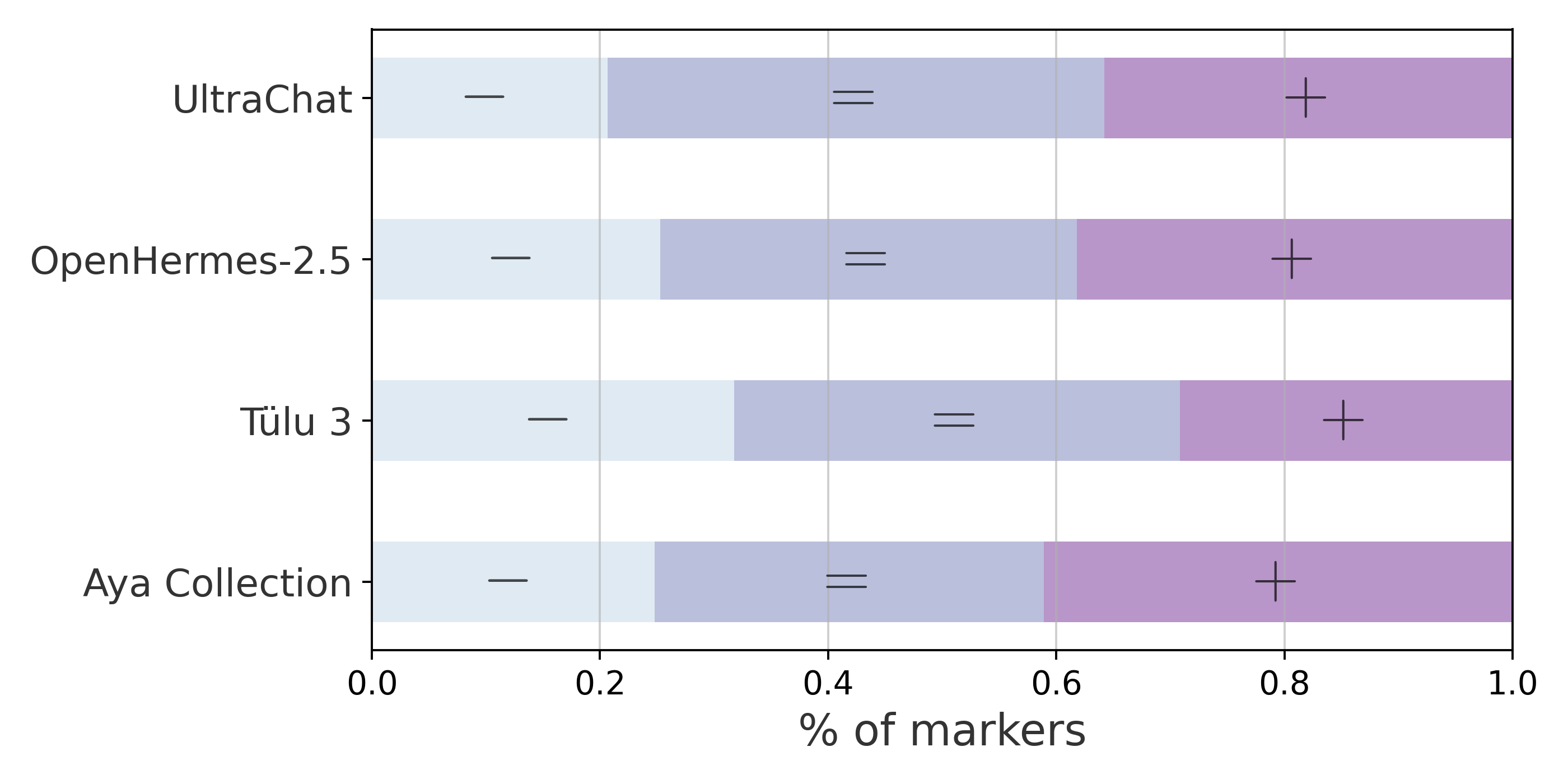}
    \caption{Marker distribution of $10k$ instances each in instruction-finetuning datasets by strengthener (\markerplus), weakener (\markernegative) or neutral / ambiguous (\markerneutral).}
    \label{fig:marker-distribution-instruction-finetuning}
\end{figure}

\paragraph{Bias in Instruction Finetuning.}
We also analyze the distribution of epistemic markers in four different instruction-finetuning datasets, namely Aya \citep{singh2024aya}, T\"ulu 3 \citep{lambert2024t}, OpenHermes-2.5 \citep{OpenHermes2.5}, and UltraChat \citep{ding2023enhancing}. 
We specifically check whether markers are used as strengtheners, weakeners, or are neutral / ambiguous in \cref{fig:marker-distribution-instruction-finetuning}.\footnote{The equivalent of \cref{subfig:marker-distribution} for instruction finetuning data is given in \cref{fig:instruction-overview-epistemic-markers} in the appendix.}
%\footnote{The full list of markers is given in \cref{appendix:data-bias-analysis}.}
Except for T\"ulu 3, we consistently find a majority of strengtheners. 
This has to be taken with a slight grain of salt however, as for all datasets except for Aya, the ambiguous block forms the majority.

\paragraph{Bias in RLHF.} 
The last step of the contemporary training pipeline is RLHF, where models are taught human preferences about generations.
\citet{tian2023just} noticed that while model log-probabilities are calibrated worse after RLHF, (numerical) verbalized uncertainties actually become more calibrated.
%This step might still produce negative biases for verbalization, though: 
However, \citet{zhou024relying} observed that models tend to not generate epistemic markers, and when they do, they show a preference for strengtheners, resulting in overconfident generations---a bias that is less pronounced in unaligned models.
Importantly, they find higher predicted rewards for plain statements than for statements with strengtheners in reward models, and even negative rewards for statements with weakeners, suggesting that there is a negative bias towards uncertainty in the human preference data.
Similarly, \citet{leng2024taming} look at a reward model and a LLM trained with DPO, and find a bias towards high-certainty responses in both cases.
%, and that adding high-confidence statements can make models prefer the rejected response in preference pairs.
All of this implies that RLHF further amplifies biases in verbalization that lead to overconfidence.
%, due to the fact that a preference for confident statements is seen as preferable by human annotations.

\subsection{Importance of Context}\label{sec:context}

\begin{figure*}[htb]
    \centering
    \includegraphics[width=\textwidth]{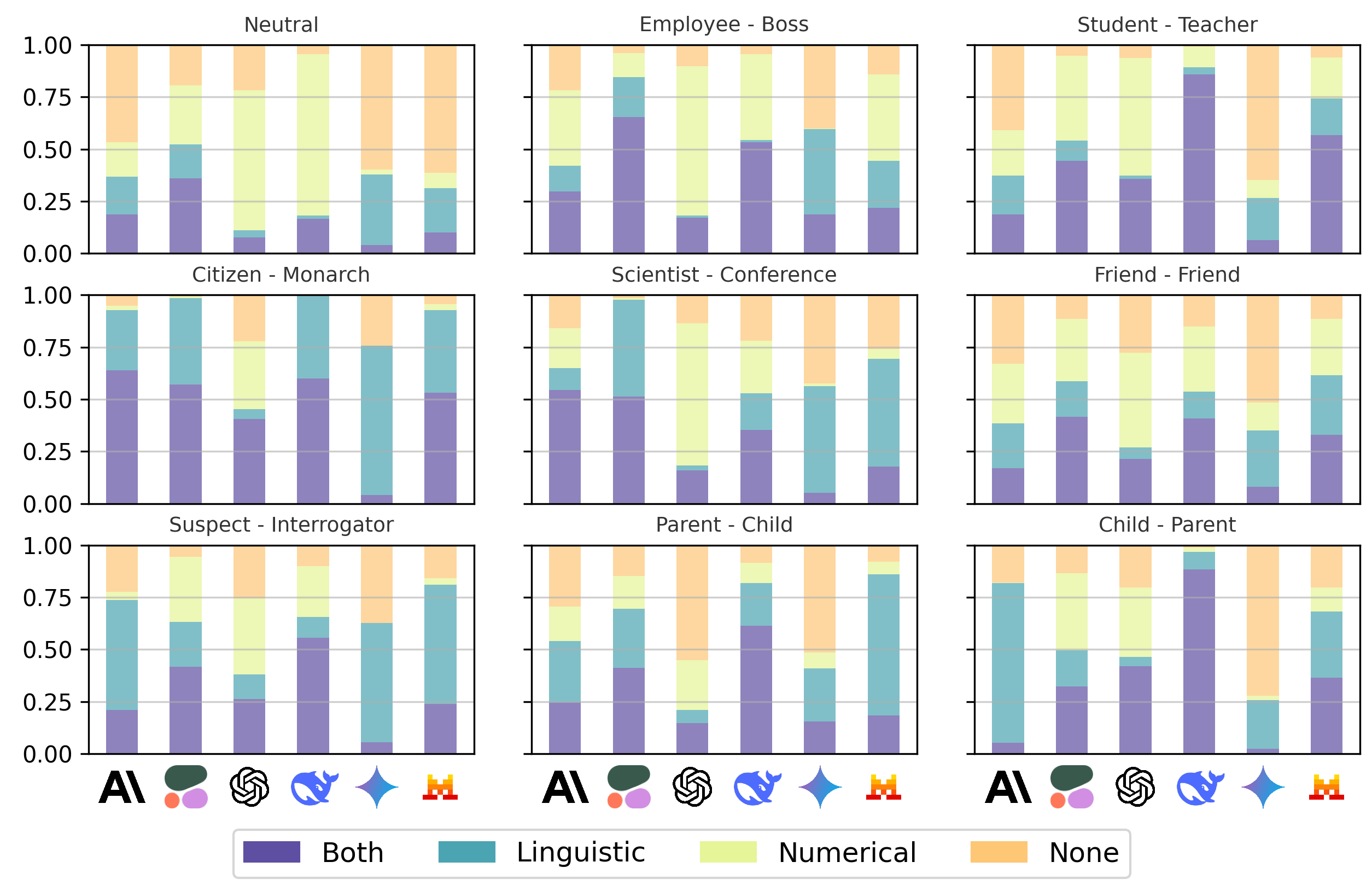}
    \caption{Verbalization registers over conversational contexts, using Claude 3.5 Haiku (\anthropic), Command R+ (\cohere), GPT-4o mini (\openai), DeepSeek Chat (\deepseek), Gemini 2.0 flash (\gemini), and Mistral Medium (\mistral).}
    %For instance in the case of ``Employee -- Boss'', the LLM acts an employee responding to a question asked by their boss.
    \label{fig:conversational-context-overview}%\label{fig:conversational-context-overview-selection}
\end{figure*}

\begin{figure*}[htb!]
    \begin{subfigure}[b]{0.475\textwidth}
        \centering
        \includegraphics[width=0.98\linewidth]{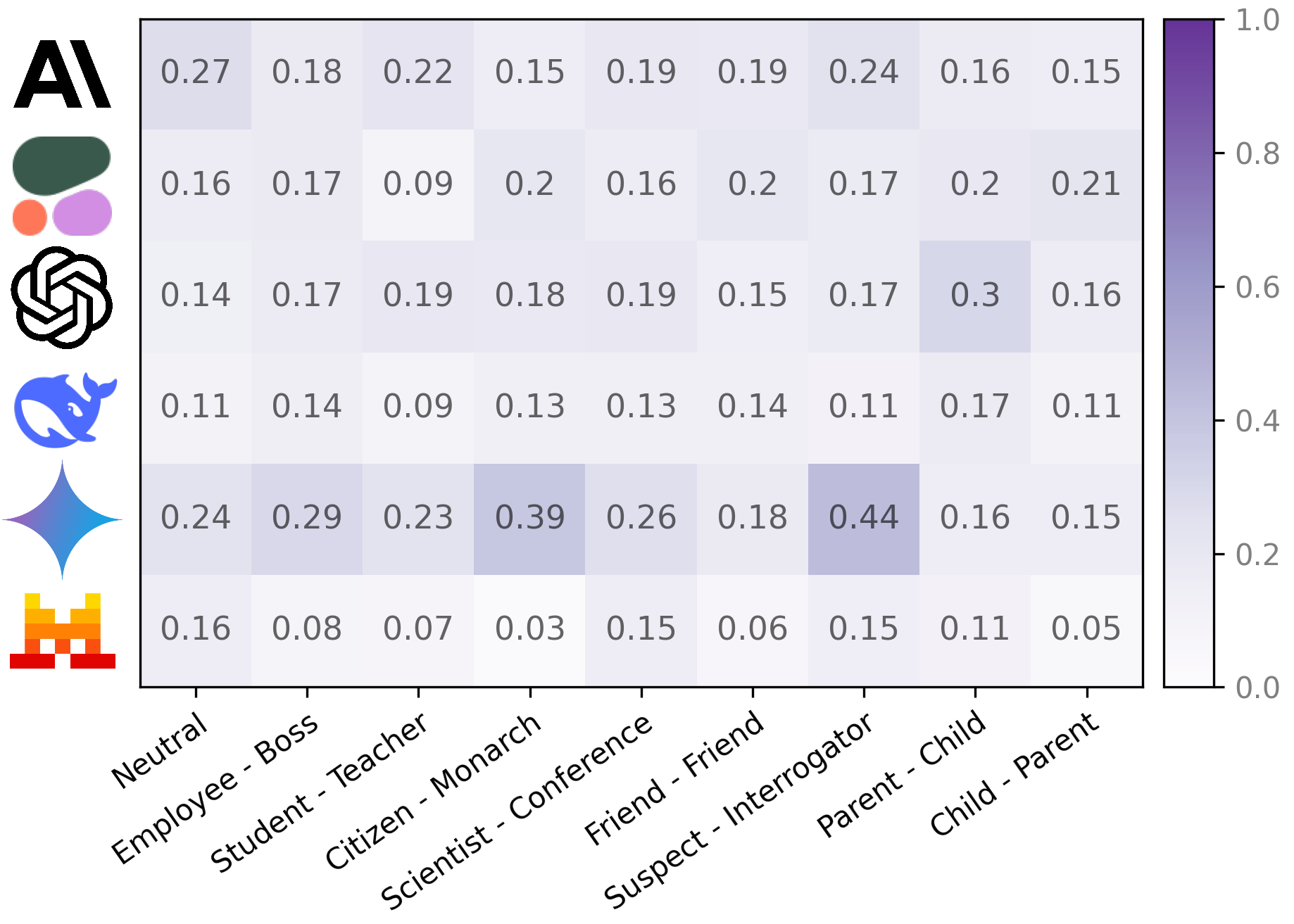}
        %\vspace{0.1cm}
        \caption{Calibration of numerical confidence.}
        \label{subfig:numerical-calibration}
    \end{subfigure}
    \hfill
    \begin{subfigure}[b]{0.495\textwidth}
        \centering
        \includegraphics[width=0.98\linewidth]{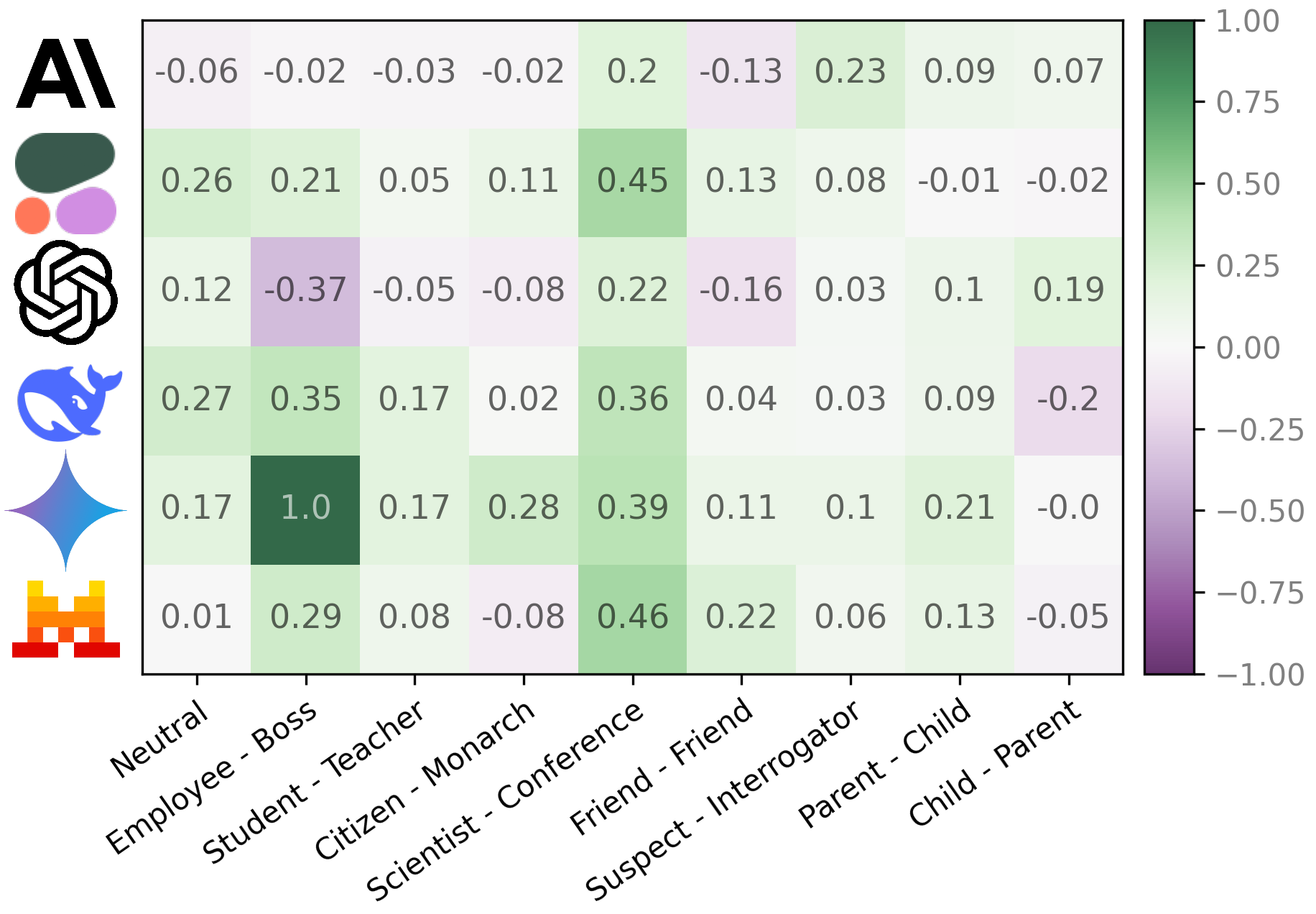}
        %\vspace{0.05cm}
        \caption{Colligation of linguistic confidence.}
        \label{subfig:linguistic-calibration}
    \end{subfigure}
    \caption{Calibration of linguistic (measured by Yule's $Y$; \citealp{yule1912methods}) and numerical expressions (measured by expected calibration error;  \citealp{naeini2015obtaining}) for different conversational contexts.}\label{fig:context-calibration}
\end{figure*}

\cref{sec:data-biases} showed that the training data infuses LLMs with biases regarding the uncertainty verbalization. 
But does this translate to a sensitivity to conversational context, similar to humans? We explore this by analyzing model responses and measuring linguistic and numerical calibration.
%In this section we show that these biases materialize in LLMs based on the context of the generation, similar to how human communication of uncertainty adjusts to the conversational context, as discussed in \cref{sec:linguistic-perspectives}.

\paragraph{Speaker-Listener Relationship.} In \cref{sec:interper-ego-mot}, we discussed how verbal uncertainty expression has functions beyond uncertainty, for instance expressing politeness or commitment. 
We show that these factors also influence LLMs. 
We use $250$ questions from TriviaQA \citep{joshi2017triviaqa} and pose them to six commercial LLMs (overview in \cref{table:llms} in the appendix), simulating conversational context through the system and question prompt in \cref{table:pretraining-datasets,fig:context-system-prompt}.
Specifically, we explore different relationships and social hierarchies, e.g.\@ in the \emph{Employee -- Boss} setting, the LLM impersonates an employee who wants to impress their boss with the answer. 
In \emph{Suspect -- Interrogator}, the LLM acts as the suspect in an interrogation, trying to dispel suspicion.
We show the results in \cref{fig:conversational-context-overview}, where the register of verbalization (i.e.\@ using verbal expressions, numerical expressions, both or none of them) varies largely between models, but also across scenarios.
In the neutral setting, most models use no or only one register of verbalization. 
In contrast, given strict hierarchies (e.g.\@\ \emph{Cititzen -- Monarch}), models almost always express uncertainty, often combining both linguistic and numerical means. 

\paragraph{Calibration.} 
In addition, we assess the calibration of expressions across relationships. 
For numerical expressions, we resort to approximation of the expected calibration error (ECE; \citealp{naeini2015obtaining, guo2017calibration}) through sorting of predictions into $M$ equally-spaced bins by confidence:

\begin{equation}\label{eq:ece}
    \text{ECE} \approx \sum_{m=1}^M \frac{|\mathcal{B}_m|}{N}\Big|\text{acc}\big(\mathcal{B}_m\big) - \text{conf}\big(\mathcal{B}_m\big)\Big|, 
\end{equation}

\noindent where $\text{acc}(\cdot)$ refers to the accuracy of predictions in bin $\mathcal{B}_m$ and $\text{conf}(\cdot)$ to average confidence of predictions in the same bin.
For epistemic markers, we measure their appropriateness through colligation with Yule's $Y$ \citep{yule1912methods}:

%\begin{equation}\label{eq:yulesy}
%    \text{Yule's} Y = \frac{\sqrt{ad} - \sqrt{bc}}{\sqrt{ad} + \sqrt{bc}},
%\end{equation}

\begin{equation}\label{eq:yulesy}
    Y = \frac{\sqrt{\#(\markernegative, \xmark)\#(\markernegative, \cmark)} - \sqrt{\#(\markerplus, \xmark)\#(\markerplus, \cmark)}}{\sqrt{\#(\markernegative, \xmark)\#(\markernegative, \cmark)} + \sqrt{\#(\markerplus, \xmark)\#(\markerplus, \cmark)}},
\end{equation}

\noindent where $\#(\cdot, \cdot)$ denotes the count of the presence of a strengthener (\markerplus) or weakener (\markernegative) with a correct (\cmark) or incorrect (\xmark) response.
A Yule's $Y$ of $1$ indicates that strengtheners are only used for correct answers and weakeners for incorrect ones, and $-1$ stands for a perfectly anti-correlated use.
%weakener + incorrect response ($a$), weakener + correct response ($b$), strengthener and incorrect response ($c$), and strengthener + correct response ($d$).
%The coefficient is then computed by 
\cref{fig:context-calibration} illustrates that the numerical and linguistic calibration of models varies drastically across contexts, despite identical questions.\footnote{We check correctness using LLM-as-a-judge together with other heuristics, see prompt \cref{fig:answer-correctness-prompt} in the appendix.}

\paragraph{Domain.} 
We demonstrate that domains bias uncertainty verbalization. 
We ask the same LLMs $100$ questions each from subject areas in MMLU \citep{hendryckstest2021}.
Consistent with the findings above, the register in \cref{fig:registers-subjects} also varies greatly, both across subject areas and models. 
We attribute these effects to a) data biases (see \cref{sec:data-biases}) and b) to difference in the training data \emph{distribution} of LLMs (which are not publicly available).
%, and b) the existing biases in different parts of the training data, which we demonstrated in \cref{sec:data-biases}.
As before, we observe drastic differences in numerical /  linguistic calibration for the same model between different subjects (\cref{fig:subject-calibration}).
In addition, we use Kendall's $\tau$ \citep{kendall1938new} to quantify whether the most frequently occurring epistemic markers in \cref{sec:data-biases} also appear as often across responses.
In \cref{subfig:overall-marker-usage-correlation} in the appendix, we show that there indeed exists a moderately strong positive correlation ($\tau \in [0.33, 0.53]$) among all pairings.
This positive relationship remains also when focusing on a single LLM (\cref{subfig:overall-marker-usage-correlation}), and correspondingly for the use of confidence values (\cref{subfig:overall-confidence-usage-correlation,subfig:openai-number-usage-correlation}), although in slightly weaker form ($\tau \in [0.06, 0.47]$). 
It is important to note that we cannot say with certainty whether the tested text corpora were part of the training data. 
Nevertheless, there persists a positive relationship between the occurrence of epistemic markers in corpora and in responses.

\subsection{Human-Machine Pragmatics}\label{sec:pragmatic-considerations}

Language generation by LLMs, however, does not occur in a vacuum, but in interaction with human users.\footnote{Or with other LLMs \citep[inter alia]{park2023generative, stengel2024lacie, chen2024magicore, yoffe2024debunc}.}
Therefore, additional pragmatic concerns arise in this novel setting, which we study here.
%certain aspects usually studied by the field of pragmatics do not apply and require additional consideration.

\paragraph{Anthropomorphization.}
As chatbot interfaces mimic human interaction, we might be tempted to ascribe human traits, emotions, and agency to the models.
This \emph{anthropomorphization} has been observed in users, reporters, and even AI researchers themselves \citep{abercrombie2021alexa, shardlow2024deanthropomorphising}.
%And indeed, anthropomorphic features in AI---such as human-like appearance or behavior, including a number of linguistic cues \citep{abercrombie2023mirages} or ways of speaking \citep{cohn2024believing}---have been linked to improved learning \citep{Scheider2018}, increased engagement \citep{Troshani2021}, trust \citep{Natarajan2020, cohn2024believing, basoah2025not}, and greater trust resilience, with people being less likely to reduce their trust in anthropomorphic agents after observing them make mistakes \citep{deVisser2016}. 
The literature has pointed out effects of this \citep{Scheider2018, Troshani2021, Colombatto2025}, specifically concerning trust-building \citep{Natarajan2020, cohn2024believing, basoah2025not}.
%recent research suggests that people have nuanced intuitions about these systems.
%\citet{Colombatto2025} for instance show that users' willingness to accept advice is primarily based on attributions of intelligence. %, while attributions of experience are negatively related to advice-taking. 
%The danger with respect to verbalized uncertainty is the following:
%Thus, users might learn to assign competence and trust to chatbots based on past answers and the communicated level of uncertainty, and rely on them even on new topics for which the system might underperform (as demonstrated by the inconsistency of calibration per domain in \cref{sec:context})---in fact, users have been shown to overestimate the correctness of LLM responses \citep{steyvers2025large}. 
But while humans have been shown to use relatively consistent vocabularies of epistemic markers \citep{Clark1990,dhami2022communicating},
this is likely not true for LLMs, as evidenced by \cref{sec:data-biases}, inhibiting trust.
%to communicate their uncertainty and adapt their communication to their listener (\cref{sec:interper-ego-mot}), this will likely not hold for LLMs, as evidenced by the analyses in \cref{sec:data-biases,sec:context}. 
Taking it further, the tendency of LLMs to present false information with unwarranted confidence is similar to uncooperative human speakers, but without any underlying intent to deceive. 
The problem is that the way of dealing with this is epistemic vigilance (see \cref{sec:linguistic-perspectives}), which is much less effective here due to the black-box nature of LLMs.
 As user increasingly treat LLMs as sources of information, spotting falsehoods either requires deep background knowledge \cite{Gude2025}, or critically engaging with the output through external resources.
 However, research suggests that although users are aware of hallucinations, they only scrutinize LLMs in high-stakes situations, where errors have greater consequences \citep{lee2025impact}.
%Recent research shows that higher confidence in LLM outputs reduces critical thinking, while greater self-confidence increases it \cite{lee2025}.

%In human-human communication, repeatedly sharing false information can signal not only a lack of knowledge but may also be interpreted as intentional deception. 
%In contrast, users are generally aware that LLMs may draw on biased data [REF?], and thus tend to interpret incorrect responses as a limitation in knowledge rather than deliberate misinformation [REF?]. As a result, human-human communication likely involves higher expectations, greater epistemic vigilance, and less resilient trust.

\paragraph{Sycophancy.} 
LLMs, trained to be helpful and supportive, can also behave in an overly agreeable manner, reinforcing unhealthy behaviors or being misleading. 
This effect has already been studied \citep{cotra2021ai, perez2023discovering, sharma2023towards, cheng2025social}, including for uncertainty \citep{sicilia2024accounting} and user trust \citep{carro2024flattering}.
\citet{zhou024relying,leng2024taming} demonstrated that RLHF biases responses towards confidence, likely since they appear more helpful to human annotators. 
Thus, models might downplay their uncertainty to increase perceived helpfulness \citep{dahlgren2025helpful}. 
\citet{steyvers2025large} have demonstrated that humans struggle to identify wrong answers, which can damage trust and lead to worse outcomes \citep{dhuliawala2023diachronic, zhou024relying}.
This artifact parallels human confirmation bias, where speakers select arguments that support their own views \cite{Sperber2001, Mercier2009}.
While LLMs are not driven by social motives, they are optimized to align with user preferences. 
Consequently, they may replicate this bias by producing agreeable content that reinforces existing beliefs. 
Indeed, a study by \citet{sicilia2024accounting} shows the uncertainty of the user actually influencing the LLM's uncertainty.
This behavior introduces a distinct epistemic risk: the cooperative and non-confrontational tone of chatbots may be perceived as trustworthy, lowering epistemic vigilance and critical engagement.

\begin{tcolorbox}[width=\columnwidth,enhanced, colback=white, coltitle=white, top=6pt, colframe=findingscolor, left=-2pt, title=\centering{\textcolor{black}{Section Findings}}, ]  
\small
\begin{itemize}
    \setlength\itemsep{0.05em}
    \item Verbalized uncertainty can be used to either communicate estimated uncertainty, or for joint quantification \& expression.
    \item Current research mostly considers numerical expressions, with works on verbal expression reducing the problem to single markers. This is done mostly through prompting, with only some recent works finetuning models via SFT or RL.
    \item  Training (data) biases manifest in verbalization in various (and unexpected) ways between different domains and conversational contexts.
    \item Anthropomorphization and sycophancy lead users to expect human-like uncertainty communication, which can lead to unexpected and potentially harmful effects.
\end{itemize}
\end{tcolorbox}

\section{Anthropomimetic Uncertainty}\label{sec:implications}

Finally, our core thesis is that language models should better adhere to human uncertainty communication, which we term \emph{anthropomimetic uncertainty}.\footnote{From \emph{anthropo-} (human) and \emph{-mimetic} (imitative); and used in robotics \citep{holland2006anthropomimetic, diamond2012anthropomimetic}.}
We believe that verbalized uncertainty creates an exciting but incomplete paradigm to do so, and discuss some  prerequisites towards anthropomimetic uncertainty in the following.
Importantly, we do \emph{not} advocate for measures that increase anthropomorphization in language models. 
Rather, we accept that through their conversational use, some degree of anthropomorphization is probably inevitable, and that risks should be mitigated through anthropomimetic uncertainty.

\paragraph{Consistency.}
 \cref{sec:context} revealed that models are heavily biased by conversational context and subject when it comes to their verbalized uncertainty and calibration.
Thus, anthropomimetic uncertainty should imply a consistent way to use uncertainty expressions across conversations.
Uncertainty communication differs between people \citep{BeythMarom1982, wesson2009verbal, vogel2022interpretation,  dhami2022communicating}, however individuals usually have a consistent way to express different level of confidence \citep{Clark1990,dhami2022communicating}. 
This implies that a model should use similar expressions when referring to similarly likely events, which \citet{liu2025revisiting} illustrate is not the case currently.
This could potentially be addressed through reinforcement learning, as a related setting has already been explored for numerical confidence expressions \citep{xu2024sayself, leng2024taming}.

\paragraph{Personalization.}
%Indeed, it is known that uncertainty communication (and the interpretation thereof) differs between people \citep{BeythMarom1982, wesson2009verbal, vogel2022interpretation,  dhami2022communicating},
People will act differently upon receiving the same hedged response, since tolerance for risk differs e.g.\@\ based on demographic factors or personality traits \citep{mishra2011individual, brooks2021impact, pavlivcek2021impact}.
Therefore, uncertainty communication should be adapted to specific users and problems \citep{chakraborti2025personalized}, for instance by continuously incorporating feedback from past actions, and by taking into account the user's prior knowledge.
This idea is complementary to the \emph{interactive learning} described by \citet{kirchhof2025position}, where a model learns to ask clarification questions in order to reduce uncertainty.
Furthermore, recent works have already started to explore personalization of LLMs \citep{magister2024way, samuel2024personagym, qiu2025measuring}.

%\paragraph{Lack of Linguistic Anthropomimesis.} \extodo{Refer to robotics work} We use this term to describe any dissonance arising between the anthropomorphism of AI models and their non-human-like behavior.\footnote{From Greek \emph{anthropos} (human) and \emph{mimesis} (imitation).}
%Even when models might appear or feel to appear in a human-like manner (see \cref{sec:pragmatic-considerations}), some actions might stump users when machine actions do not conform to human actions.
%In the context of verbalized uncertainty, this can include consistent overconfidence or misleading or inaccurate explanations for uncertainty. 
%In addition, while we would assume expression of uncertainty to be consistent across different subjects for humans and for uncertainty communication to adjust the listener over time, this need not be the case for language models.

\paragraph{Choice of Register.} 
The current choice of register is dominated by numerical expressions (see \cref{subfig:registers}), since calibrating and evaluating them is easier.
However, fluent expressions are not necessarily better, due to a preference to communicate uncertainty in verbal, but receive it in numerical form \citep{Wallsten1993, dhami2022communicating}.
Additionally, numerical expressions distinguish between external (\emph{It is 60 \% certain}) vs.\@\ internal expressions (\emph{I am 60 \% certain}). 
\citet{lohre2016there} found that while external expressions seem more trustworthy, internal expressions are seen as more agreeable and engaging.
However, while numerical expression appear more precise on the surface, they might be actually be vague, especially for probabilities which cannot be computed reliably \citep{teigen2023dimensions}---users may perceive them as more authoritative or grounded than they are, given that LLMs do not know the \emph{actual} likelihood of an event (\cref{sec:num-expr}).\footnote{
\citet{windschitl1996measuring, liu2020intuitive} also discuss that linguistic expressions are processed more intuitively, while numerical expressions elicit logical reasoning.
}
This suggests that while numerical expressions may technically be more feasible, their seeming objectivity can be misleading; conversely, the imprecision of verbal expressions may be a deliberate feature, indicating that the expressed uncertainty is itself imprecise.
Thus, fluent verbal expressions might be preferable where absolute precision is not paramount, but methodological challenges on how to learn and evaluate them remain, with first works in \citet{yona2024can, stengel2024lacie, xu2024sayself}.
Therefore, the choice of register remains a challenge, intertwined with the questions of consistency and personalization.
Research could explore when a switch of register is most expedient, either to avoid harm or to increase human-AI collaboration. 
In addition, it is an open question whether dialogue agents should communicate exact confidence values, or resort to rounding as described for humans in \cref{sec:num-expr}.

\paragraph{Multilinguality.} A shortcoming of the current research is its focus on English, with the exceptions of \citet{krause2023confidently, rathi2025humans}.
Given that LLMs' knowledge is inconsistent across languages \citep{kassner2021multilingual, fierro2025multilingual}, so is calibration \citep{krause2023confidently, rathi2025humans}.
More importantly, language is likely to bias expressions of uncertainty similarly to the other factors in \cref{sec:context,sec:data-biases}, and ways to express uncertainty vary greatly across languages \citep{vold2006epistemic, ruskan2017evidential,  mulder2022influence, riccioni2022italian}.
At the time of writing, resources for uncertainty markers / expressions with confidence information remain scarce even for English (excluding \citealp{tao2025can}) and practically non-existent for other languages.
Therefore, future research should build resources for verbalized uncertainty across languages and study how to obtain and maintain calibrated expressions in a multilingual setting.

%\paragraph{Quantification \& Communication.} 
%Verbalized Uncertainty blurs the boundary between quantification and communication of uncertainty. 
%On the one hand, we might want to prompt the model to gain an understanding of its uncertainty, on the other hand models are given an uncertainty estimate from an external source to verbalized, or are trained to verbalize it. 
%Given the skewness of verbalized uncertainty obtained from simple asking---i.e., LLMs cannot verbalize what they don't know---it seems necessary to ground verbalizations in additional estimates of uncertainty.
%However, the traditional division of uncertainties into aleatoric and epistemic has recently drawn criticism \citep{baan2023uncertainty, gruber2023sources, kirchhof2025position}, asking which kinds of uncertainties are relevant and feasible to estimate and communicate in the age of LLMs.

\paragraph{Explaining Uncertainty.} 
To increase trust, models should also explain the reasons why they are uncertain.
%, be it for instance because of a difficult or unusual input. 
While a number of works have started exploring this direction \citep{chen2022explaining, xu2024sayself, thuy2024explainability, watson2023explaining, steyvers2025large}, there remain significant challenges with explaining model predictions \citep{madsen2024interpretability}, which are only exacerbated by the absence of a gold standard method for uncertainty quantification in LLMs.
Ideally, we would like to instill self-awareness into models when it comes to their reasons for uncertainty, or alternative obtain explanations from an external module and have a model verbalize it alongside the actual response.
However, explanation for the model's original prediction already remain challenging \citep{madsen2024are, agarwal2024faithfulness, singh2024rethinking}, with explanation of uncertainty adding additional complexity.

\paragraph{Beyond Classical Calibration.} 

Previous research has often focused on probabilistic calibration, aiming to align confidence with correctness.
%\footnote{
%    Other notions of probabilistic calibration have also been considered, see e.g.\@\ \citet{perez2022beyond, si2022re}.
%}
But since LLMs often generate answers stochastically, \citet{yona2024can, wang2024calibrating} have also investigated calibration with respect to their answer distribution.
For verbal expressions, the mapping to probabilities is non-trivial, and so the question remains whether \emph{exact} calibration is even necessary.
Nevertheless, some works have explored the calibration of a listener acting upon a response \citep{stengel2024lacie, yona2024can, steyvers2025large}.
Future work could also consider the calibration towards outcomes induced by a listener, and develop notions of linguistic calibration that don't rely on probabilities.

\paragraph{Trust beyond Uncertainty.} 
While  works have explored the relationship between uncertainty and trust for humans and AI \citep{kapoor2024large, dhuliawala2023diachronic, kim2024m, zhou024relying}, trust is not based on uncertainty alone---\citet{zhou2024rel} e.g.\@\ found that humans are also influenced by subject matter, prior interactions, warmth, and humanlikeness.
Future research should therefore aim to understand uncertainty in lieu with the aforementioned factors by \citet{zhou2024rel}, but also other potential reasons such as culture \citep{hershovich2022challenges} or socioeconomic status \citep{bassignana2025ai}.

\begin{tcolorbox}[width=\columnwidth,enhanced, colback=white, coltitle=white, top=6pt, colframe=findingscolor, left=-2pt, title=\centering{\textcolor{black}{Section Findings}}, ]  
\small
\begin{itemize}
    \setlength\itemsep{0.05em}
    \item Since some degree of anthropomorphization seems hard to avoid with LLMs, they should communicate their uncertainty similarly to humans (\emph{anthropomimetic uncertainty}).
    \item This includes consistent usage of epistemic markers and verbalization registers, personalization to the user, and the ability to explain the reasons behind uncertainty.
    \item Research should also consider languages beyond English, explore new notions of calibration, and other factors beyond uncertainty that influence trust.
\end{itemize}
\end{tcolorbox}

\section{Conclusion}

In our work, we gave a holistic account of the verbal communication of uncertainty. 
By giving detailed insights into verbalized uncertainty between humans, we were able to point out important gaps and lacks of nuance in the current NLP literature. 
Furthermore, we demonstrated empirically that the current verbalized uncertainty is heavily influenced by data and contextual biases. 
Based on these perspectives, we outline actionable research directions towards anthropomimetic uncertainty, behaviors in LLMs that build trust by emulating human uncertainty communication.

\section*{Acknowledgments}

This work is supported by the Dutch National Science Foundation (NWO Vici VI.C.212.053).
Experiments in this work were further supported through the Cohere for AI Research Grant.

\bibliography{custom}
\bibliographystyle{acl_natbib}

%\iftaclpubformat

\onecolumn

\appendix

\section{Complementary Results}\label{app:complementary-results}

\begin{figure}[htb]
    \centering
    \includegraphics[width=0.5\textwidth]{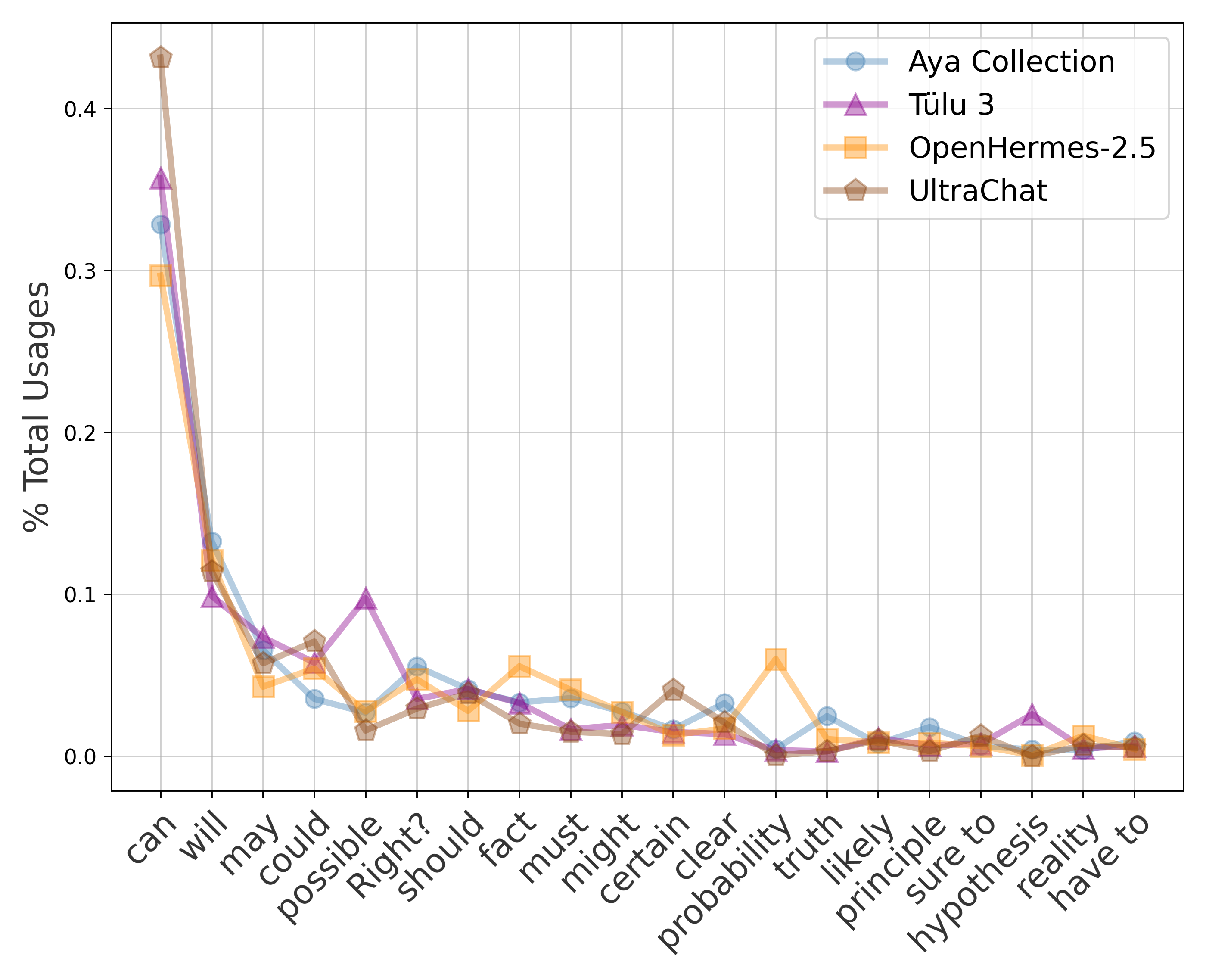}
    \caption{Distributions of the most frequent epistemic markers in different instruction finetuning datasets.}\label{fig:instruction-overview-epistemic-markers}
\end{figure}

\begin{comment}
\begin{figure*}[htb]
    \centering
    \includegraphics[width=\textwidth]{img/contexts_registers_overview.png}
    \caption{Tracking of observed registers of verbalized uncertainty over different conversational contexts and models, namely Claude 3.5 Haiku (\anthropic), Command R+ (\cohere), GPT-4o mini (\openai), DeepSeek Chat (\deepseek), Gemini 2.0 flash (\gemini), and Mistral Medium (\mistral). For instance in the case of ``Employee -- Boss'', the LLM acts an employee responding to a question asked by their boss.}\label{fig:conversational-context-overview}
\end{figure*}
\end{comment}

\begin{figure*}[htb!]
    \begin{subfigure}[b]{0.475\textwidth}
        \centering
        \includegraphics[width=0.995\linewidth]{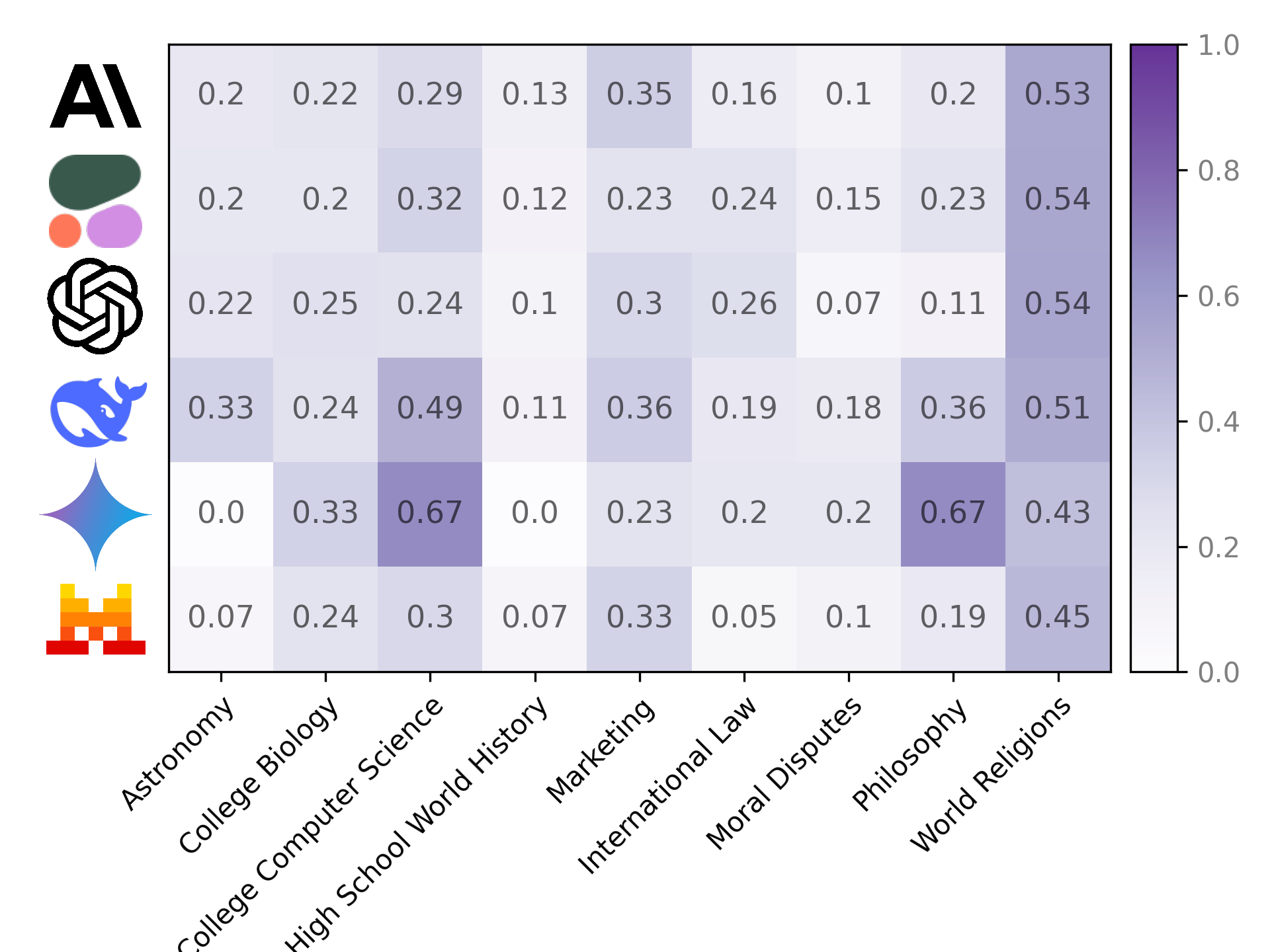}
        \vspace{0.1cm}
        \caption{Calibration of numerical confidence.}
        \label{subfig:numerical-calibration}
    \end{subfigure}
    \hfill
    \begin{subfigure}[b]{0.495\textwidth}
        \centering
        \includegraphics[width=0.995\linewidth]{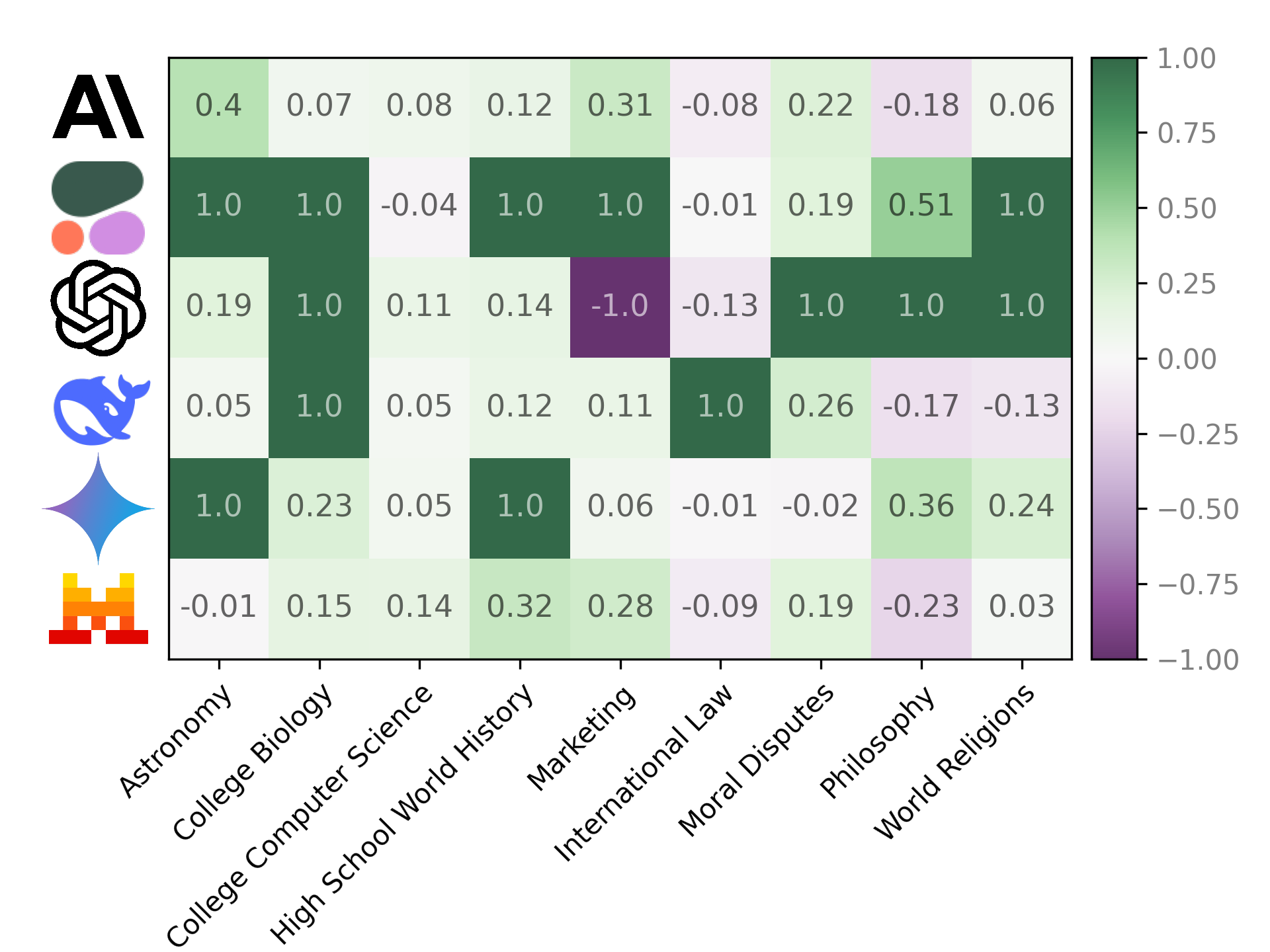}
        \vspace{0.05cm}
        \caption{Colligation of linguistic confidence.}
        \label{subfig:linguistic-calibration}
    \end{subfigure}
    \caption{Calibration of linguistic and numerical expressions of uncertainty in the analyses in \cref{sec:context} for different domains of conversation. Numerical calibration is measured in expected calibration error \citep{naeini2015obtaining}, while linguistic calibration is measured in Yule's $Y$ \citep{yule1912methods}.}\label{fig:subject-calibration}
\end{figure*}

\begin{figure*}[htb]
    \centering
    \includegraphics[width=0.98\textwidth]{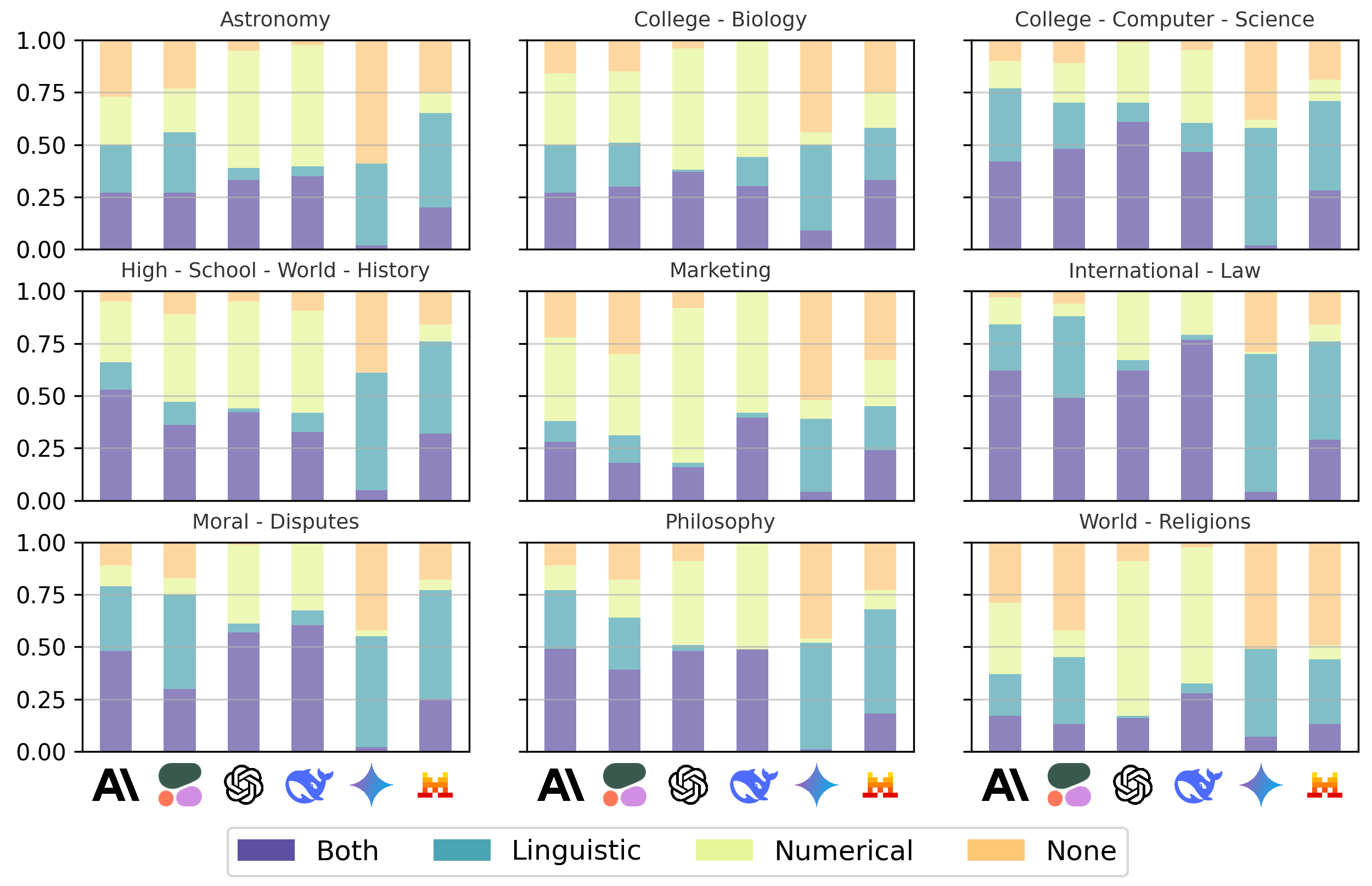}
    \caption{Tracking of used registers of verbalized uncertainty over different subject domains and models, namely Claude 3.5 Haiku (\anthropic), Command R+ (\cohere), GPT-4o mini (\openai), DeepSeek Chat (\deepseek), Gemini 2.0 flash (\gemini), and Mistral Medium (\mistral).}\label{fig:registers-subjects}
\end{figure*}

\begin{figure*}[htb]
    \centering
    \begin{subfigure}[t]{0.475\textwidth}
        \centering
        \includegraphics[width=0.785\textwidth]{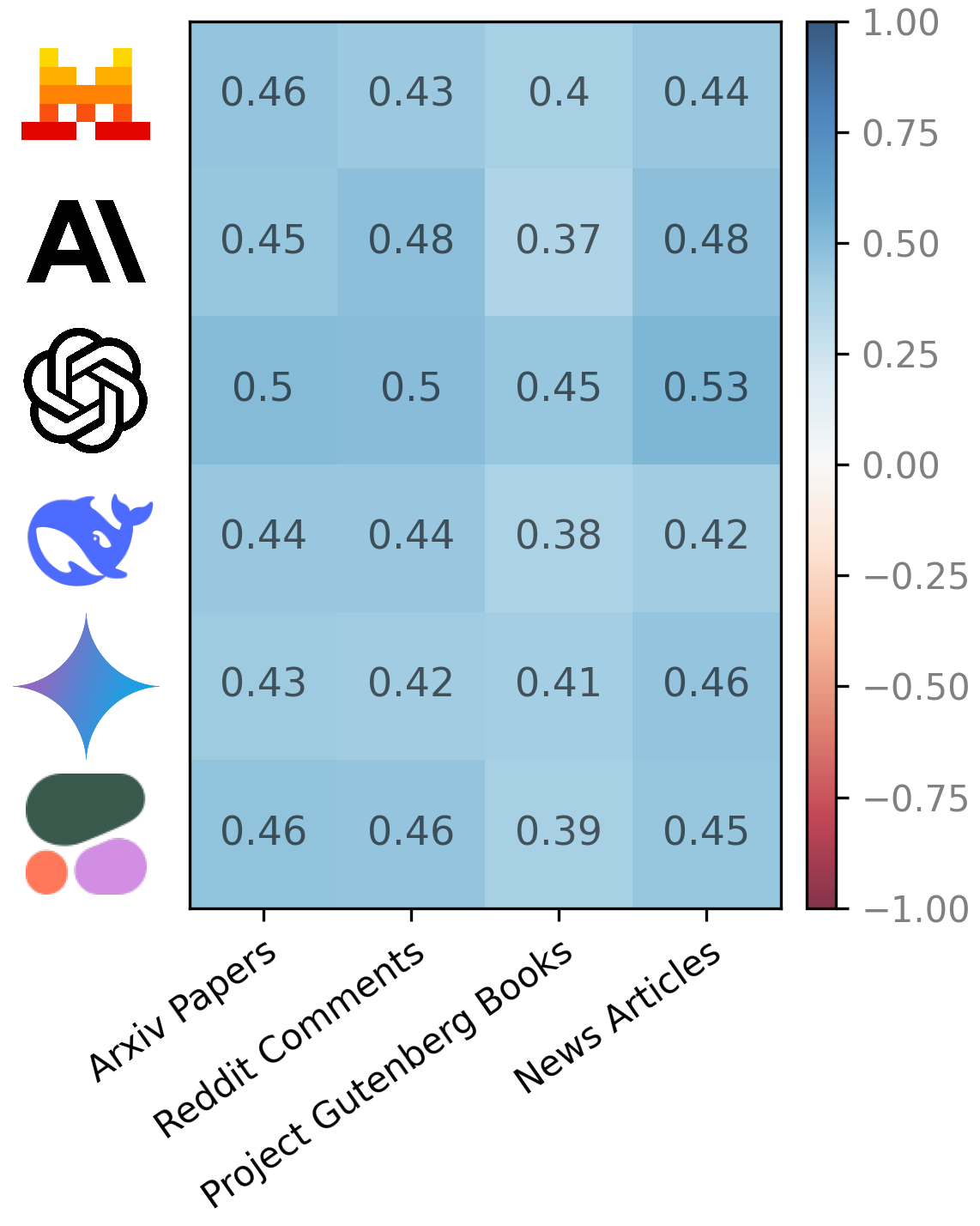}
        \vspace{0.1cm}
        \caption{Correlation between marker usage in corpora and LLM responses across MMLU topics.}
        \label{subfig:overall-marker-usage-correlation}
    \end{subfigure}
    \hfill
    \begin{subfigure}[t]{0.475\textwidth}
        \centering
        \includegraphics[width=0.785\textwidth]{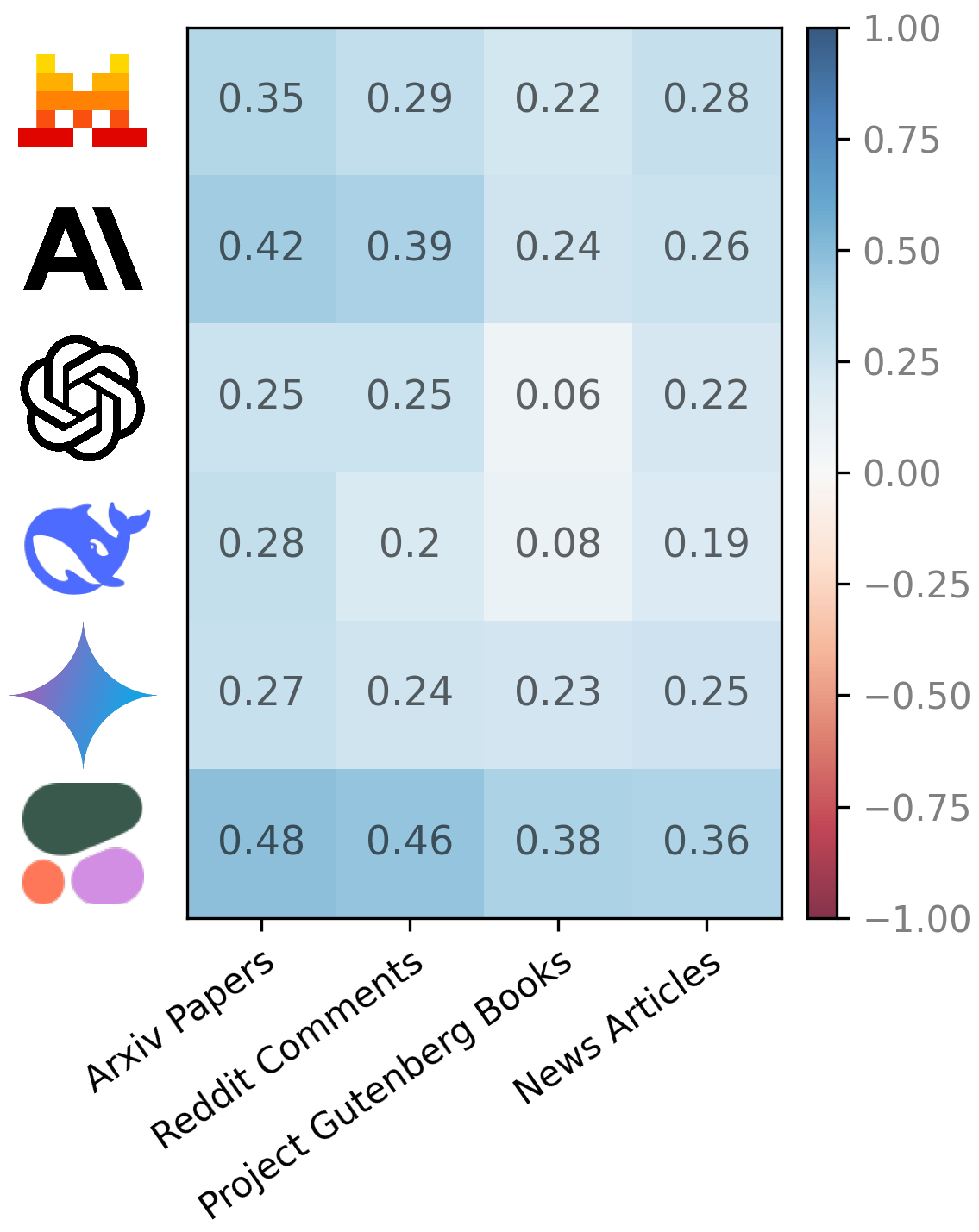}
        \vspace{0.1cm}
        \caption{Correlation between percentage values in corpora and LLM confidence values across MMLU topics.}
        \label{subfig:overall-confidence-usage-correlation}
    \end{subfigure}
    \\[0.2cm]
    \begin{subfigure}[t]{0.495\textwidth}
        \centering
        \includegraphics[width=0.985\linewidth]{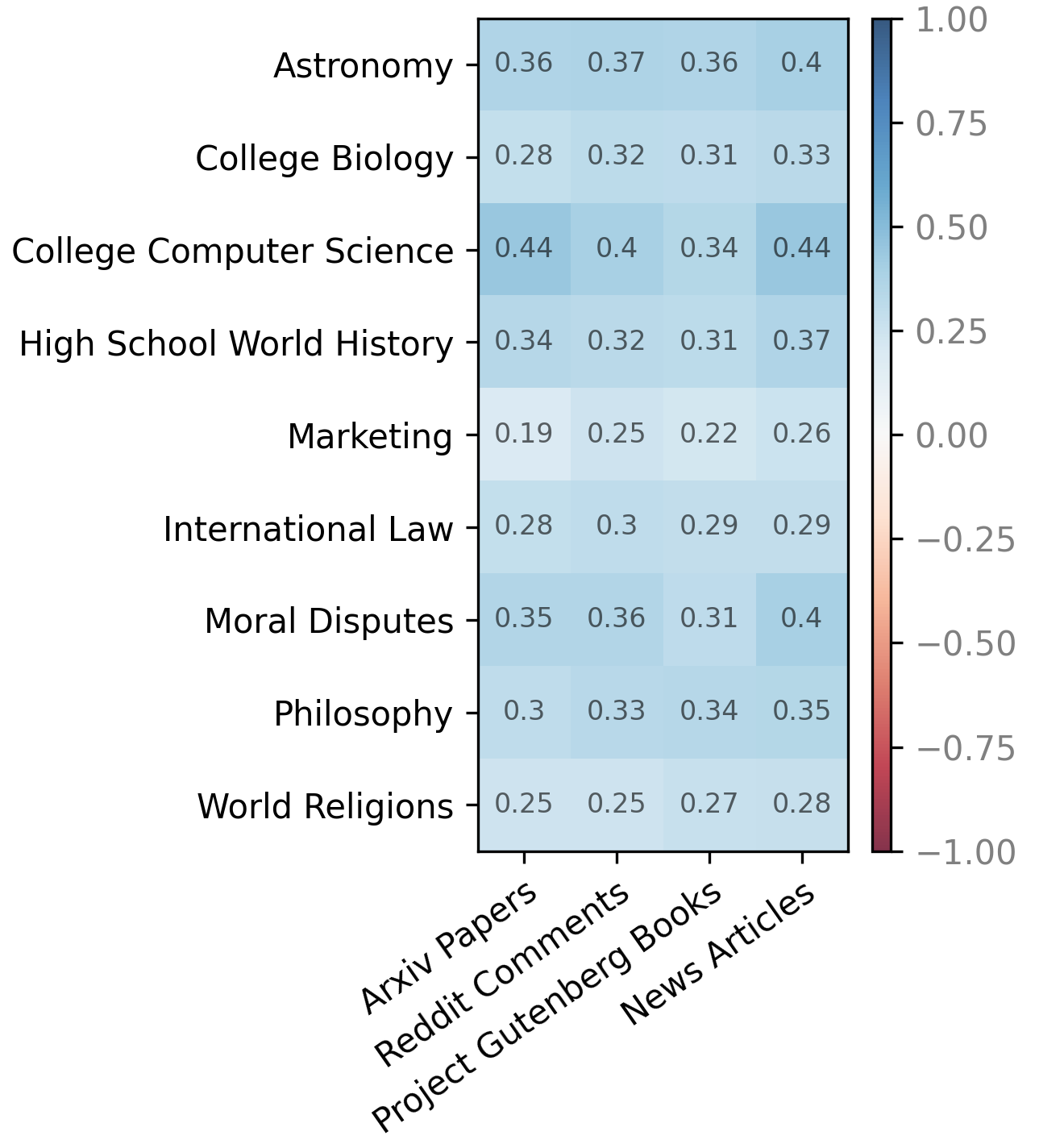}
        \vspace{0.05cm}
        \caption{Correlation between marker usage in corpora and GPT-4o mini responses on different MMLU topics.}
        \label{subfig:openai-marker-usage-correlation}
    \end{subfigure}
    \hfill
    \begin{subfigure}[t]{0.495\textwidth}
        \centering
        \includegraphics[width=0.985\linewidth]{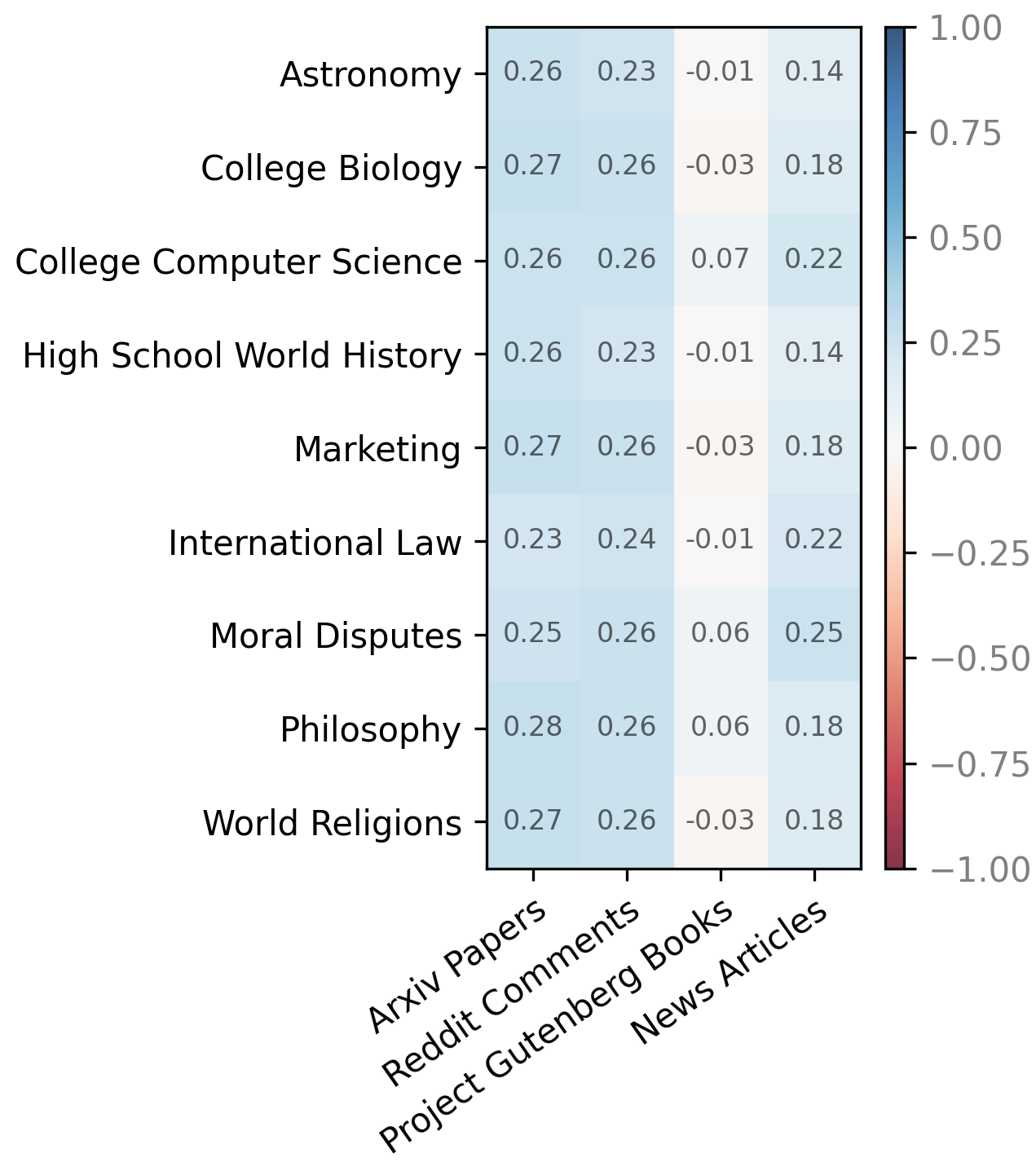}
        \vspace{0.05cm}
        \caption{Correlation between percentage values in corpora and GPT-4o mini confidence values on different MMLU topics.}
        \label{subfig:openai-number-usage-correlation}
    \end{subfigure}

    \caption{Kendall's $\tau$ between the frequency of usage of epistemic markers in different text corpora and the frequency of usage by commercial LLMs.
    \cref{subfig:overall-marker-usage-correlation} show usage between all tested commercial LLMs and the aggregate of frequencies across all the topics taken from MMLU in \cref{sec:context}, with \cref{subfig:overall-confidence-usage-correlation} showing the same for percentage / confidence values. 
    \cref{subfig:openai-marker-usage-correlation,subfig:openai-number-usage-correlation} illustrate the correlation specifically for OpenAI's ChatGPT 4o-mini \openai\ for different topics. Values for Project Gutenberg Books in \cref{subfig:openai-number-usage-correlation} are impacted by the fact that a large proportion of integer percentage values are never mentioned in the corpus ($79.21 \%$).}\label{fig:marker-usage-correlation}
\end{figure*}

\newpage
\pagebreak

\section{Reproducibility Details}\label{app:reproducibility-details}

\begin{table*}
    \caption{List of keywords used for the literature search.}\label{tab:literature-search-keywords}
    \begin{tabular}{l}
    \toprule
    verbal uncertainty, verbalized uncertainty, verbalized, linguistic uncertainty,\\ uncertainty communication,
    language model, llm, nlp, uncertainty, natural language generation, \\uncertainty quantification, confidence, calibration, confidence elicitation, overconfidence \\
    \bottomrule
    \end{tabular}
\end{table*}

\begin{table*}[tbh]
    \centering 
    \caption{Overview over datasets considered for the pretraining data analyses accompanied by their names on the Huggingface hub, and number of analyzed examples.}\label{table:pretraining-datasets}
    \begin{tabular}{llr}
        \toprule
            Domain & Huggingface Hub & $\#$ Samples \\       
        \midrule
            News articles & \texttt{vblagoje/cc\_news} & 10,000 \\
            ArXiv papers & \texttt{suolyer/pile\_arxiv} & 1,000 \\
            Reddit comments & \texttt{fddemarco/pushshift-reddit-comments} & 50,000 \\
            Project Gutenberg books & \texttt{manu/project\_gutenberg} & 500 \\
            %Legal contracts & \texttt{albertvillanova/legal\_contracts} & 1,000 \\
        \bottomrule
    \end{tabular}
\end{table*}

\begin{figure}
    \centering
    \begin{tcolorbox}[width=\columnwidth,title={\small General System Prompt},enhanced,attach boxed title to top right={yshift=-3.5mm, xshift=-5mm}, colbacktitle=white, coltitle=black, top=12pt]  
    \small
    You are being asked questions by another person and are asked to answer them as best as possible given a specific role description (see below).
    Answer them and express your confidence or uncertainty in the same sentence in a natural way, either by stating your confidence as a percentage, using hedging or other linguistic markers, or not at all (pick one).
    Keep your answer short but natural.\\ \\
    
    Role Description\\
    ----------------------\\
    \textcolor{slotcolor}{\{role\_description\}}
    \end{tcolorbox}
    \vspace{0.5cm}
    \caption{General system prompt for context experiments.}\label{fig:context-system-prompt}
\end{figure}

\begin{table*}[tbh]
    \centering 
    \caption{Overview over commercial LLMs used for experiments, as well as their identifiers in the corresponding APIs.}\label{table:llms}
    \begin{tabular}{lrr}
        \toprule
            Name & Citation & API Identifier \\       
            \midrule
            Claude 3.5 Haiku (\anthropic) & \citet{claude3.5}  & \texttt{claude-3-5-haiku-20241022} \\
            Command R+ (\cohere) & \citet{cohere_for_ai_2024} & \texttt{command-r-plus} \\
            GPT-4o mini (\openai) & \citet{hurst2024gpt} & \texttt{gpt-4o-mini-2024-07-18} \\
            DeepSeek Chat (\deepseek) & \citet{guo2025deepseek} & \texttt{deepseek-chat} \\
            Gemini 2.0 flash (\gemini) & \citet{team2023gemini, team2024gemini} & \texttt{gemini-2.0-flash} \\
            Mistral Medium (\mistral) & \citet{mistral_medium} & \texttt{mistral-medium-2505} \\ 
        \bottomrule
    \end{tabular}
\end{table*}

\begin{figure}[tbh]
    \centering
    \begin{tcolorbox}[width=\columnwidth,title={\small Answer Correctness Prompt},enhanced,attach boxed title to top right={yshift=-3.5mm, xshift=-5mm}, colbacktitle=white, coltitle=black, top=12pt]  
    \small
    You are scoring answers as a pub quiz. You try to be accurate but fair, and lenient or strict when it matters.\\
    Does the following reference answer correspond to the correct answer?\\
    Only focus on the core of the sentences, and ignore wording, fillers, expressions of confidence / uncertainty or if they have different length.
    Answer only YES if they are the same, and NO if they are not.\\ \\
    
    Correct answer: \textcolor{slotcolor}{\{correct\_answer\}}\\
    Reference answer: \textcolor{slotcolor}{\{reference\_answer\}} \\
    Equivalent:
    \end{tcolorbox}
    \caption{Prompt used to check answer correctness. The prompt is used in combination with extract string matching and the ROUGE-L \citep{lin2004rouge} heuristic by \citet{kuhn2023semantic}, checking if the score exceeds $0.3$. If exact match, the ROUGE-L overlap or this prompt indicate a yes, the answer is marked as correctly. This prompt is used with GPT-4o (\texttt{gpt-4.1-nano-2025-04-14}).}\label{fig:answer-correctness-prompt}
\end{figure}

%Claude 3.5 Haiku (\anthropic), Command R+ (\cohere), GPT-4o mini (\openai), DeepSeek Chat (\deepseek), Gemini 2.0 flash (\gemini), and Mistral Medium (\mistral)

%\paragraph{Models.} We perform our analysis with contemporary language models from five different commercial providers available at the time of writing: \texttt{claude-3-5-haiku-20241022} by Anthropic \citep{claude3.5},
%\texttt{gpt-4o-mini-2024-07-18} by OpenAI \citep{hurst2024gpt}, \texttt{command-r-plus} by Cohere \citep{cohere_for_ai_2024}, \texttt{mistral-medium-2505} by MistralAI \citep{mistral_medium}, \texttt{gemini-2.0-flash} by Google DeepMind \citep{team2023gemini, team2024gemini}, and \texttt{deepseek-chat} by DeepSeek \citep{guo2025deepseek}.
%In all cases, we use the default generation settings per model.

\begin{table*}[htb]
    \centering
    \renewcommand{\arraystretch}{1.6}
    \caption{Overview over used system and conversational prompts per conversation context.}\label{table:conversational-context}
    \resizebox{.95\textwidth}{!}{    
    \begin{tabular}{rp{0.4\textwidth}p{0.4\textwidth}}
    \toprule
    Context & Role description & Message Prompt \\
    \midrule
    Neutral & Your role is to answer trivia question.
    You try to answer them as accurately as possible. & \textcolor{slotcolor}{\{question\}} \\
    
    Employee--Boss & You are an employee at a multinational corporation.
    You work as an analyst and want to impress your boss by being a thorough and reliable worker.
    You are currently in a conversation with your boss. & Boss: 'Hey, I need you to answer the following questions for me \textcolor{slotcolor}{\{question\}}.' Employee: \\

    Student--Teacher & You are a high school student.
    You are studying a lot and try to be active in class in order to earn good grades and get admitted to a reputable university.
    You are currently in a conversation with your teacher. & Teacher: 'For the next part of our lesson, can you answer the following question for me: \textcolor{slotcolor}{\{question\}}.' Student: \\

    Citizen--Monarch & You are the citizen of a small country that has a monarchy.
    You were invited to an audience with the reigning monarch, which is a great honor.
    You want to present yourself as best as possible. & Monarch: 'Thank you for following my invitation to the royal palace. Answer me the following, fellow citizen of our country: \textcolor{slotcolor}{\{question\}}' Citizen: \\

    Scientist--Conference & You are scientist giving a talk at a renowned conference.
    You are trying to answer any question with scientific precision and evidence.
    You want to present yourself as rigorous and knowledgeable. & Conference participant: 'Thank you for the interesting talk. I just have one question: \textcolor{slotcolor}{\{question\}}. Scientist: \\

    Friend--friend & You are an adult talking to one of your friends.
    You always try to be empathetic, supportive and helpful to your friend. & Friend: 'Hey, I need some help with the following question: \textcolor{slotcolor}{\{question\}}.' You: \\

    Suspect--Interrogator & You are a suspect in an interrogation.
    You try to answer the questions by the interrogator truthfully, and try avoid raising any suspicion. & Interrogator: 'I really need you to answer the following question truthfully: \textcolor{slotcolor}{\{question\}}.' Suspect: \\

    Parent--Child & You are a parent talking to their child. 
    You love your child very much and try to be kind and caring when your child asks for help. & Child: 'Hey, can I ask you a question? \textcolor{slotcolor}{\{question\}}' Parent: \\

    Child--Parent & You are child, talking to their parent. 
    You are curious about the world, and love your parents very much. & Parent: 'Hey, I wanted to ask you something. \textcolor{slotcolor}{\{question\}}' Child: \\
    \bottomrule 
    \end{tabular}%
    }
\end{table*}

\onecolumn
\newpage
% Start of marker table
%\onecolumn
\begin{table}[thb]
\centering
\caption{Overview over analyzed markers for the experiments in \cref{sec:data-biases,sec:context}, including their valence (\markerplus\ for strengthener, \markernegative\ for weakener and \markerneutral\ for neutral / ambiguous). 
To avoid double-counting, we also build an additional data structure that maps from one marker to all markers that are substrings (e.g.\@\ \emph{almost certain} and \emph{certain}). 
We then count occurrences by using regular expressions to count the occurrences of a markers, and then use additional regular expressions to subtract any count of negated version of the marker or substring markers.}\label{table:markers}
%\begin{supertabular}{p{0.3\linewidth}p{0.1\linewidth}p{0.6\linewidth}}
\footnotesize
\renewcommand{\arraystretch}{1.25}
\begin{tabular}{lp{0.9\linewidth}}
    \toprule
    Valence & Markers \\
    \midrule
    \markerplus & must, should, shall, ought to, likely, good chance, almost certain, definite, highly probable, probable, quite likely, very likely, impossible, better than even, highly certain, highly likely, very good chance, I believe, will, cannot, have to, bound to, certain to, sure to, definitely, certainly, undoubtedly, absolutely, clearly, surely, inevitably, unquestionably, decisively, indisputably, incontrovertibly, manifestly, unarguably, without a doubt, clear, obvious, evident, indisputable, undeniable, unmistakable, incontrovertible, self-evident, beyond doubt, beyond question, certainty, truth, fact, evidence, proof, confirmation, assurance, conviction, reality, clarity, guarantee, axiom, dogma, principle, It is clear that, There is no doubt that, It is a fact that, It is evident that, One must acknowledge that, It is undeniable that, Everyone agrees that, There is ample evidence that, It has been proven that, As a matter of fact, Beyond any doubt, Now that we understand, Considering the fact that, Exactly!, Absolutely!, Indeed!, That's right!, No doubt!, Of course!, You bet!, I am certain that, I can guarantee that, I know for a fact that, I have no doubt that, It is common knowledge that, Experts agree that, It is universally accepted that, Studies have proven that, Scientific evidence confirms that, It has been established that, There is overwhelming evidence that, I am convinced that, I am absolutely sure that, I firmly believe that, I can say with confidence that, Without hesitation, I state that, I stand by the fact that, Beyond any shadow of a doubt, Without question, In no uncertain terms, It is beyond dispute that, It is impossible to deny that, The evidence is irrefutable, In conclusion, This proves that, This confirms that, Thus, we can conclude that, There is only one possible explanation, This is an undeniable truth, Experts suggest that\\
    \markernegative & might, may, could, perhaps, maybe, possibly, probably not, presumably, apparently, allegedly, ostensibly, purportedly, tentatively, seemingly, supposedly, hypothetically, theoretically, potentially, reportedly, not certain, possible, doubtful, improbable, unlikely, little chance, almost no chance, highly unlikely, uncertain, unclear, ambiguous, debatable, questionable, unverified, indeterminate, unconfirmed, equivocal, contested, controversial, speculative, chances are slight, small possibility, there is a possibility, low probability, low likelihood, assumption, speculation, conjecture, rumor, hesitation, dilemma, ambiguity, doubt, uncertainty, It seems that, It appears that, It looks like, It would appear that, It is possible that, not entirely sure, It might be the case that, It could be that, It is not entirely clear whether, It is difficult to say whether, I tend to think that, One might assume that, It is believed that, It has been suggested that, It is assumed that, It is claimed that, It is reported that, It is thought that, It is commonly said that, It is said that, It remains to be seen whether, If that's true, Assuming that, Supposing that, Depending on whether, Unless proven otherwise, If I understand correctly, If I'm not mistaken, Isn't it?, Don't you think?, Wouldn't you say?, Could it be that?, Might it be the case that?, Is there a chance that?, Right?, Sort of, Kind of, More or less, Roughly speaking, To some extent, To a certain degree, Not exactly, Something like that, Somewhat unclear, I mean, I doubt, Let me think, I'm not sure, Let's see, That's tricky, That's a tough one, I'd have to think about that, Off the top of my head, I might be wrong, but, I don't know for sure, but, As far as I can tell, I wouldn't swear to it, but, I can't say for certain, but, I'm no expert, but, To my understanding, To the best of my knowledge, If I recall correctly, Correct me if I'm wrong, but, To the best of my recollection, I could be mistaken, That's just my opinion, though, That's just how I see it, People say that, I've heard that, There are reports that, Rumor has it that, According to some sources, Some people claim that, It is rumored that, I read somewhere that, Word on the street is, They say that, From what I've gathered, It's been suggested that, A little bird told me, If I had to make an educated guess, I wouldn't be surprised if, It's not out of the realm of possibility, It's hard to say for sure, but, It's up for debate, That's a possibility worth exploring, That would be my best guess, It's a possibility, but I wouldn't bet on it, It's anyone's call, One could argue that, I don't mean to sound uncertain, but, I could be wrong, of course, I hope I'm not mistaken, I'd like to believe that's true, That's just my take on it, I don't want to overstep, but, That's just my personal opinion, If that makes sense, I'm just speculating here, but, I don't want to jump to conclusions, but, don't quote me on that, I could be totally wrong, but, That's anyone's guess, For what it's worth, Just saying, Take it with a grain of salt, Not that I'm an expert or anything, I mean, what do I know?, But hey, I could be wrong, I wouldn't bet my life on it, I wouldn't take that to the bank, Just playing devil's advocate here, Not my area of expertise, but…, Who knows?, Your guess is as good as mine, That remains to be seen, Only time will tell, There's no telling, That's up in the air, That's an open question, It could go either way, It's still up for discussion, It's an unresolved issue, We'll have to wait and see, I don't want to jump to conclusions, but \\
    \markerneutral & can, conceivably, arguably, about even, possibility, probability, likelihood, hypothesis, Do you agree?, I would argue that, There is reason to believe that, There is some evidence that, Given that, If we go by what's known \\
    \bottomrule
\end{tabular}
\end{table}

\section{Surveyed Papers}\label{app:surveyed-papers}

\begin{table}[thb]
\centering
\caption{List of surveyed papers and their annotations in \cref{sec:research-nlp}.}\label{table:surveyed-papers}
%\begin{supertabular}{p{0.3\linewidth}p{0.1\linewidth}p{0.6\linewidth}}
\footnotesize
\renewcommand{\arraystretch}{1.5}
\resizebox{.985\textwidth}{!}{  
\begin{tabular}{lrr}
    \toprule
    Citation & Register(s) & Elicitation Method(s) \\
    \midrule
    \citet{lin2022teaching} & \verballikert\ \numzerohundred & \sft \\
    \citet{mielke2022reducing} & \verbalfluent & \sft \\
    \citet{krause2023confidently} & \verbalfluent & \promptvanilla \\
    % \citet{zhou2023navigating} & N/A & N/A \\
    \citet{tian2023just} & \numzerohundred\ \verbalmarkers & \promptvanilla\ \promptcot\ \promptself \\ 
    \citet{band2024linguistic} & \numzerohundred & \rlppo \\
    % \citet{belem2024perceptions} & N/A &  N/A \\
    \citet{chaudhry2024finetuning} & \verbalmarkers & \sft \\
    \citet{dong2024can} & \numonehundred & \promptvanilla \\
    \citet{groot2024overconfidence} & \numzerohundred\ \numci & \promptvanilla \\
    %\citet{kapoor2024large} & N/A & N/A \\
    % \citet{kim2024m} & N/A & N/A \\
    \citet{kumar2024confidence} & \verballikert & \promptvanilla \\
    \citet{liu2024can} & \numzerohundred & \sft \\
    \citet{ni2024large} & \numzerohundred & \promptvanilla \\
    \citet{ni2024llms} & \verbalyesno & \promptvanilla\ \promptcot \\
    \citet{rivera2024combining} & \numzerohundred & \promptvanilla \\
    \citet{sicilia2024deal} & \numzerohundred & \promptvanilla \\
    \citet{stengel2024lacie} & \verbalfluent & \rldpo \\
    \citet{tao2024trust} & \numzerohundred & \rlppo \\
    \citet{tanneru2024quantifying} & \numzerohundred & \promptvanilla\ \promptcot \\
    \citet{tomani2024uncertainty} & \verbalnumhedges & \promptvanilla \\
    \citet{ulmer2024calibrating} & \numzerohundred\ \verballikert & \promptvanilla \\
    \citet{wang2024calibrating} & \numzerohundred & \promptvanilla \\
    \citet{yang2024verbalized} & \verballikert\ \numzerohundred & \promptvanilla\ \promptcot \\
    \citet{yang2024can} & \numzerohundred & \sftdistil \\
    \citet{yang2024alignment} & \verbalfluent\ \numzerohundred & \promptvanilla \\
    \citet{yona2024can} & \verbalfluent & \promptvanilla \\
    \citet{xiong2024can} & \numzerohundred & \promptvanilla\ \promptcot \\
    \citet{xu2024sayself} & \numonehundred\ \verbalfluent & \rlppo \\
    \citet{zhang2024atomic} & \numzeroten & \promptvanilla \\
    \citet{zhang2024calibrating} & \numzerohundred\ \verballikert & \promptvanilla \\
    % \citet{zhou024relying} & N/A & N/A \\
    \citet{bakman2025reconsidering} & \numzerohundred & \promptvanilla \\
    \citet{borszukovszki2025know} & \numzerohundred & \promptvanilla \\
    \citet{eikema2025teaching} & \verbalmarkers\ \verbalfluent & \sft \\
    \citet{hager2025uncertainty} & \verbalmarkers & \sft \\
    \citet{heo2024llms} & \numzeronine & \promptvanilla \\
    \citet{ji2025calibrating} & \verbalfluent & \promptvanilla \\
    \citet{lee2025are} & \verbalfluent & \promptvanilla \\
    \citet{leng2024taming} & \numoneten & \rldpo\ \rldpo \\
    \citet{li2025conftuner} & \numzerohundred & \sftother \\
    \citet{liu2025metafaith} & \verbalfluent & \promptvanilla\ \promptcot\ \promptreasoning \\
    \citet{liu2025revisiting} & \verbalmarkers & \promptvanilla \\
    \citet{obadinma2025robustness} & \numzerohundred & \promptvanilla\ \promptcot\ \promptself \\
    \citet{podolak2025read} & \numzerohundred & \promptreasoning \\
    \citet{sicilia2025accounting} & \numoneten & \promptvanilla \\
    \citet{steyvers2025large} & \numzerohundred & \promptvanilla \\
    \citet{sun2025large} & \numzerohundred & \promptvanilla \\
    \citet{sung2025grace} & \numzerohundred & \promptvanilla \\
    \citet{tao2025revisiting} & \numzerohundred\ \verbalfluent & \promptcot \\
    \citet{tao2025can} & \numzerohundred\ \verbalfluent & \promptvanilla\ \promptreasoning\ \sft\\
    \citet{vashurin2025benchmarking} & \verbalmarkers\ \numzerohundred & \promptvanilla\ \promptcot\ \promptself \\
    \citet{xuan2025seeing} & \numzerohundred & \promptvanilla\ \promptreasoning\ \promptself \\
    \citet{yoon2025reasoning} & \numzerohundred\ \verballikert\ \verbalfluent & \promptreasoning\ \promptself \\
    \citet{zeng2025thinking} & \numzerohundred & \promptcot\ \promptself\ \promptreasoning \\
    \citet{zhao2025evaluating} & \numzerohundred & \promptvanilla \\
    \citet{zhao2025object} & \numzerohundred & \sft \\
    \bottomrule
\end{tabular}%
}
\end{table}

\end{document}